\documentclass[conference]{IEEEtran}

\usepackage[numbers]{natbib}
\usepackage{xurl}
\usepackage{amsmath}
\usepackage{verbatim}
\usepackage{amsfonts}
\usepackage{multirow}
\usepackage{booktabs}
\usepackage{amssymb}
\usepackage{pifont}% http://ctan.org/pkg/pifont
\newcommand{\cmark}{\ding{51}}%
\newcommand{\xmark}{\ding{55}}%
\usepackage{soul,xcolor}
\usepackage{floatrow}
\usepackage{soul}
\usepackage[tableposition=top]{caption}
\usepackage{caption}
\usepackage{siunitx}
\usepackage{adjustbox}
\usepackage{lscape}
\usepackage{rotating}
\usepackage{cleveref}
\usepackage[normalem]{ulem}
\usepackage{xcolor,colortbl}
\usepackage{color, colortbl}
\usepackage[short]{optidef}
\usepackage{subcaption}
\usepackage{xurl}
\usepackage{algorithm, algpseudocode}
\usepackage{balance}
\usepackage{tabularx}
\usepackage{tcolorbox}
\usepackage{url}
\usepackage{listings}
\newtcolorbox[auto counter, number within=section]{myfigure}[2][]{%
  float*=htbp, width=\textwidth, colframe=black, colback=white, 
  boxrule=0.5mm, title=#2,#1}

\ifCLASSINFOpdf

\else

\fi

\begin{document}
\title{LLMs are One-Shot URL Classifiers and Explainers}

\author{\IEEEauthorblockN{Fariza Rashid\IEEEauthorrefmark{1},
Nishavi Ranaweera\IEEEauthorrefmark{2},
Ben Doyle\IEEEauthorrefmark{3},
Suranga Seneviratne\IEEEauthorrefmark{1}}
\IEEEauthorblockA{\IEEEauthorrefmark{1}
University of Sydney \IEEEauthorrefmark{2}University of New South Wales \IEEEauthorrefmark{3}Thales Australia}
%\IEEEauthorblockA{\IEEEauthorrefmark{2}University of New South Wales}
%\IEEEauthorblockA{\IEEEauthorrefmark{3}Thales Australia}
\IEEEauthorblockA{\{fariza.rashid, suranga.seneviratne\}@sydney.edu.au}
\IEEEauthorblockA{r.tharushika\_nishavi@unsw.edu.au, ben.doyle@thalesgroup.com.au}}

\maketitle

\begin{abstract}
%\boldmath
Malicious URL classification represents a crucial aspect of cyber security. Although existing work comprises numerous machine learning and deep learning-based URL classification models, most suffer from generalisation and domain-adaptation issues arising from the lack of representative training datasets. Furthermore, these models fail to provide explanations for a given URL classification in natural human language. In this work, we investigate and demonstrate the use of Large Language Models (LLMs) to address this issue. Specifically, we propose an LLM-based one-shot learning framework that uses Chain-of-Thought (CoT) reasoning to predict whether a given URL is benign or phishing. We evaluate our framework using three URL datasets and five state-of-the-art LLMs and show that one-shot LLM prompting indeed provides performances close to supervised models, with GPT 4-Turbo being the best model, followed by Claude 3 Opus. We conduct a quantitative analysis of the LLM explanations and show that most of the explanations provided by LLMs align with the post-hoc explanations of the supervised classifiers, and the explanations have high readability, coherency, and informativeness.

\end{abstract}

\section{Introduction}
\label{Sec:Introduction}

Phishing attacks over emails and other similar channels such as SMS remain a significant concern in cybersecurity. More often than not, a security breach can be traced back to an initial entry caused by a phishing attack. A recent report sheds light on this issue, noting a 40\% increase in phishing attacks between 2022 and 2023, with over 709 million attempts by users to access phishing links blocked by Kaspersky~\cite{kaspersky2024}. With attackers' ability to reach a wide audience in a relatively short time and using short-lived campaigns, defending against phishing remains a highly challenging task that requires multi-faceted solutions. 

% A recent report from Kaspersky sheds light on this, reporting a 40\% increase in phishing attacks between 2022 and 2023 and over 709 million attempts by users to access phishing links which were blocked by Kaspersky~\cite{kaspersky2024}
%\textcolor{blue}{With attacker’s ability to reach a wide audience in a relatively short time and using short-lived campaigns, defending}
%\textcolor{red}{TODO - See whether you can add some stats related to phishing with citations./ things to look for - some global phishing stats, a reference for security breaches tracing back to phishing}

Typical blacklisting and whitelisting methods for phishing detection are ineffective in large-scale and dynamic environments. Typically, there is a delay between identifying a phishing campaign and blacklisting the related URLs, by which time the campaign may have already succeeded~\cite{oest2020phishtime}. Consequently, machine learning-based solutions have been proposed for phishing URL detection. These methods extract features from emails and URLs, such as word counts, word lengths, and character distribution, as well as external data like WHOIS and IP information~\cite{jain2019machine, alani2022phishnot, chin2018phishlimiter}. The extracted features are then used to train machine learning and deep learning models to predict whether a given email or URL is benign or phishing. %\textcolor{red}{TODO - Add citations.}

%\textcolor{blue}{Typical blacklisting and whitelisting methods for phishing detection are ineffective in large-scale and dynamic environments. Typically, there is a delay between identifying a phishing campaign and blacklisting the related URLs, by which time the campaign may have already succeeded. Consequently, machine learning-based solutions have been proposed for phishing URL detection. These methods extract features from emails and URLs, such as word counts, word lengths, and character distribution, as well as external data like WHOIS and IP information. The extracted features are then used to train machine learning and deep learning models to predict whether a given email or URL is benign or phishing.}

Among these solutions, some research focuses on predicting phishing links based solely on their URL patterns. This approach has several advantages: it does not require visiting the web page to gather the necessary features~\cite{sahoo2017malicious}, it is resilient to evasion techniques like cloaking~\cite{zhang2021crawlphish}, and it is more suitable for detecting zero-day phishing campaigns for which prior information is unavailable. The intuition behind these methods is that phishing URLs often display distinct patterns, such as random string sequences and text patterns that mimic known URLs, along with unique artefacts from domain generation algorithms. In contrast, benign URLs tend to maintain consistent and recognisable characteristics~\cite{nowroozi2022adversarial}. %\textcolor{red}{it might be good if we can include a citation for this - or a version of this argument.} %\textcolor{red}{TODO - It is still shaky. See whether you can imporve it and add necessary references.}

Although current state-of-the-art URL classifiers such as URLNet~\cite{le2018urlnet} and URLTran~\cite{maneriker2021urltran} achieve high detection accuracies on test URLs from the same source as their training data, their performance drops significantly on test URLs from diverse sources. This decline is mainly due to high false positive rates, indicating a poor understanding of benign URL indicators~\cite{rashid2024phishing}. One reason for this issue is the inherent biases in URL datasets, which are collected at different vantage points of various networks. Moreover, data from one organisation will be biased towards URLs frequently visited by its employees. Sharing data between organisations could potentially address these biases; however, privacy and commercial concerns make it impractical. 

Furthermore, existing URL classification models are often black boxes, lacking explainability, which is crucial for user awareness and training. That is, given a model's prediction for whether a specific URL is benign or phishing, it is useful to provide users with brief explanations and warnings to help them make informed decisions~\cite{covidxai, greco2023explaining, Bender_Horn_Loewenstein_Roberts_2024}. Some examples include Outlook~\cite{outlookwarning} and Google~\cite{googlewarning} which provide users with warnings when they accidentally try to access a suspicious link. Althobaiti et al.~\cite{althobaiti2021don} further states that in phishing attack detection it is necessary to empower the end-users with more knowledge about a specific situation where a URL is potentially unsafe. Most existing tools that address this are either aimed at experts in the cyber security domain or only provide users with binary decisions about the URL. Such binary advice fails to gain the trust of the users when the decisions have a high false positive rate. For this reason it is important to provide users with simple and easy-to-understand explanations for a certain URL classification.

Recent advances in large language models (LLMs) offer potential solutions to the generalisability and explainability problems of phishing URL classifiers. Trained on vast amounts of data, LLMs excel in natural language text generation and reasoning and are useful for various tasks. Works such as~\cite{jin2023time, arora2022ask, loukas2023making} have demonstrated LLMs' predictive performance, which has been enhanced by their recently discovered \textit{in-context learning capabilities} such as few-shot learning~\cite{kojima2022large, brown2020language}. Given that LLMs are trained on a significant portion of internet data, they likely have some knowledge of benign URLs in a broader context. This knowledge can be combined with in-context learning to build URL classifiers, which is the focus of this paper.

To this end, in this paper, we propose an LLM-based phishing URL detection framework that integrates both technological and human components of an ideal defence mechanism against phishing campaigns. Our framework leverages LLMs’ one-shot learning capabilities for phishing URL classification and provides explanations for each classification. By leveraging on the success of \textit{Chain-of-Thought (CoT)} reasoning~\cite{wei2022chain}, we prompt the LLM to consider the benign and phishing characteristics of a given URL and then make a prediction, including self-reasoning. In our prompt, we include one URL, an explanation of its benign and malicious characteristics, and a prediction as illustrated in Figure 1. Our results show this approach achieves detection accuracies comparable to supervised deep learning methods trained on large URL datasets. Additionally, the LLM’s reasoning provides natural language explanations that enhance user awareness of benign and phishing URL characteristics, addressing the limitations of existing classifiers that only offer a prediction. %\textcolor{red}{Add a prompt and the output example figure here - pick a really good example.}

% \begin{tcolorbox}[colback=blue!5!white, colframe=blue!50!black, title=An Example Prompt and the LLM output, boxrule=0.5mm, width=\columnwidth, halign=left, coltitle=white, fonttitle=\bfseries]
% \textbf{Consider whether the URL seems benign or phishing in 40 words, end sentence with a prediction, the last word being benign/phishing.\\Q: \url{http://scholar.google.com.pk/citations?user=IkvxoFIAAAAJ&hl=en} \\
% A: The subdomain of this URL is a well-known and reputable internet entity, Google Scholar. Therefore, this URL seems benign. \\ 
% Q: \url{https://www.rt.com/tags/football/}}\\

% \texttt{A: The URL directs to a known news website, RT, specifically its section on football. Given its straightforward path and reputable source, this URL appears unthreatening. Prediction: benign.}

% \end{tcolorbox}
% \noindent\begin{minipage}{.5\textwidth}
% \captionof{figure}{}\label{fewshot_class_framework}
% \end{minipage}

\begin{tcolorbox}[colback=gray!10!white, colframe=black, title=An Example Prompt and the LLM output]
\underline{Prompt:}
Consider whether the URL seems benign or phishing in 40 words, end sentence with a prediction, the last word being benign/phishing. \\ \vspace{-3mm}

{\bf Q:} \fontsize{8pt}{10pt}\selectfont % Adjust the font size here
\texttt{\url{http://scholar.google.com.pk/citations?user=IkvxoFIAAAAJ&hl=en}} \\ \vspace{-3mm}

\normalsize{}
%\texttt{\url{http://www.vivscreisveci.vcirveseiaveesi.ghqphy.top/uWBRvZ8quj/page1.php}} \\

{\bf A:} The subdomain of this URL is a well-known and reputable internet entity, Google Scholar. Therefore, this URL seems benign. \\ \vspace{-3mm}

{\bf Q:} \fontsize{8pt}{10pt}\selectfont % Adjust the font size here
\texttt{\url{https://www.rt.com/tags/football/}} \\ 

\normalsize{}
\underline{Output:}\\
\textit{{\bf A:} The URL directs to a known news website, RT, specifically its section on football. Given its straightforward path and reputable source, this URL appears unthreatening. {\bf \textit{Prediction: benign}}.} \\ \vspace{-3mm}

\end{tcolorbox}
\noindent\begin{minipage}{.5\textwidth}
\captionof{figure}{Example URL classification prompt and the output}\label{llm_indicator_prompt}
\end{minipage}

%[cite papers using few-shot/in-context learning for cyber security purposes]

%In our paper, we propose an LLM-based phishing URL detection framework that satisfies both the technological and human components of an ideal defence mechanism against phishing campaigns. Our framework leverages LLMs’ few-shot learning capabilities for malicious URL classification and explaining each classification. Motivated by the success of chain-of-thought reasoning, we first prompt the LLM to consider the benign and phishing characteristics of a given URL and then ask it to make a prediction. In the rest of this paper, we refer to this reasoning step as the LLMs' self-explanations for their predictions. As a few-shot example, we include in our prompt one URL, an explanation that states the benign and malicious characteristics of the URL, and finally a prediction. Our initial analysis revealed that the reasoning approach allows the LLMs to return more accurate final predictions than simply asking the LLMs to give a prediction only. 

More specifically, we make the following contributions.

\begin{itemize}
  \item We propose an LLM-based framework leveraging Chain-of-Thought reasoning and in-context one-shot learning for phishing URL classification and demonstrate that LLMs are explainable one-shot phishing URL classifiers. 
  \item We evaluate our framework over five state-of-the-art LLMs and three different phishing URL datasets and compare the performance of our framework with existing supervised URL classifiers.
  %\item We demonstrate that our one-shot approach achieves prediction performance within 0.05 - 0.12 F1 score of a supervised setting using standard URL classifiers. Out of the five LLMs we evaluated, GPT 4-Turbo was found to be the best performing, with an average F1 score of 0.94 in the one-shot setting. This is within 0.05 points of a completely supervised setting, which returns a 0.99 F1 score.%\textcolor{red}{here F1 score is not accuracy}
  \item We demonstrate that our one-shot approach achieves a prediction performance within 0.05 to 0.12 F1 score of a supervised setting using standard URL classifiers. Among the five LLMs evaluated, GPT-4 Turbo performed the best, with an average F1 score of 0.92 in the one-shot setting. This is only 0.07 points below the fully supervised setting, which achieves a 0.99 F1 score.
  
  \item We demonstrate the explainability of our classification framework in terms of correctness and overall quality of language. That is, we assess the correctness of LLM self-explanations by comparing the benign and malicious indicators identified through our framework with post-hoc explanations obtained in a supervised setting. We use the G-Eval framework~\cite{liu2023g} to assess the self-explanations in terms of readability, coherence, and informativeness.
  %Secondly, we conduct a user study to obtain a human evaluation of the same comparison.
  \item Finally, we present an extended analysis of the consistency of LLM predictions and the performance of our framework under zero-shot and few-shot settings. Our results indicate that GPT-4 Turbo achieves an average F1 score of 0.92 in both the zero-shot and five-shot settings suggesting that increasing the number of examples has little effect on prediction accuracy.

\end{itemize}

%The rest of the paper is organised as follows. In Section~\ref{Sec:Related}, we present the related work, and in Section~\ref{Sec:Framework}, we present our overall framework. We position our experiment design in Section~\ref{Sec:predacc_eval}. We show our main results in Section~\ref{Sec:Results} and extended results analysis in Section~\ref{Sec:ExtendedAnalysis}. We discuss the implications and limitations of our work in Section~\ref{Sec:Discussion}, and we conclude the paper in Section~\ref{Sec:Conclusion}.

The remainder of the paper is structured as follows: Section~\ref{Sec:Related} reviews related work, and Section~\ref{Sec:Approach} details our overall framework. Section~\ref{Sec:predacc_eval} outlines our experiment design, while Section~\ref{Sec:Results} presents our main results. An extended analysis of these results is provided in Section~\ref{Sec:ExtendedAnalysis}. We discuss the implications and limitations of our work in Section~\ref{Sec:Discussion}, and conclude the paper in Section~\ref{Sec:Conclusion}.

\section{Related Work}
\label{Sec:Related}

\subsection{Phishing URL Detection}
Automated phishing URL classification is a necessity to safeguard users from accessing phishing sites. Multiple existing solutions utilise machine learning and deep learning methods, which are trained using features extracted from the URLs, landing pages, or external sources. These features, broadly categorised as blacklist features, lexical features, host-based features, and content-based features~\cite{sahoo2017malicious}, have been used to train various types of machine learning and deep learning classification models~\cite{mourtaji2021hybrid, liu2020spwalk, rao2021heuristic}. WHOIS information such as domain age and page rank from the Alexa top domains list are also popular features used in various works such as~\cite{xiang2011cantina, geng2015combating, rao2019detection, alani2022phishnot, chin2018phishlimiter}. 

One significant disadvantage of using the above features for detecting phishing URLs is the difficulty in obtaining them in real-time. Blacklist features can be redundant for short-lived or zero-day phishing URLs, as these exploit the delay in blacklist updates. Acquiring hosting information during the training phase is often impossible for short-lived URLs, and using web-page content features risks accidental malware downloads and is vulnerable to evasion techniques~\cite{oest2020phishtime}. Consequently, some URL classifiers use only URL-based features, which eliminates reliance on third-party sources, mitigates malware risks and reduces vulnerability to evasion. Our work focuses on these URL-based phishing detectors that use either hand-crafted URL features or an embedding representation of the URL as the primary input feature. Works such as ~\cite{mamun2016detecting, sahingoz2019machine, rao2020catchphish} collected hand-crafted features like URL length, word counts and presence of special characters to train Random Forests and Support Vector Machines. In contrast, the URLNet~\cite{le2018urlnet} trains a deep learning-based classifier using word and character embeddings to represent the URL, and Maneriker et al.~\cite{maneriker2021urltran} proposed URLTran, a model which leveraged state-of-the-art transformer models for phishing detection. In comparison to URLNet, Maneriker et al. report that URLTran returns a true positive rate (TPR) relative improvement of 21.9\% at the false positive rate (FPR) of 0.01\%.

The high prediction accuracies reported in existing works were obtained by evaluating URL classifiers on test URLs collected from the same source as the training URLs. When evaluated on data collected from different sources or representing domain shift, these URL classifiers demonstrated low generalisability. Our previous work~\cite{rashid2024phishing} found that the cross-dataset performance of four URL classifiers (URLNet~\cite{le2018urlnet}, URLTran~\cite{maneriker2021urltran}, CatchPhish~\cite{rao2020catchphish}, and PhishRF~\cite{sahingoz2019machine}) falls by 10\%--30\% in comparison to their performance on test URLs collected from the same source as the training URLs. 

\textit{To this end, in this paper, we propose our one-shot framework for using LLMs as URL classifiers and demonstrate that its performance is better than the cross-data performance of URL classifiers trained in supervised settings. To the best of our understanding, this work is the first to design an LLM-based one-shot URL classification framework.}

%\textcolor{red}{I didn't chech this - It is quite long. I don't think you have to spent too much time on describing it. Try to shorten it.}

\subsection{Few-shot classification using LLMs}

Large Language Models (LLMs) have demonstrated superior performance in a variety of natural language processing tasks, including text generation, question-answering, and text classification~\cite{achiam2023gpt, touvron2023llama, narang2022pathways}. Recently, their generalisability has seen further improvements within the in-context learning paradigm where the LLM performs tasks unseen during training when prompted with task-specific instructions and a few or no (i.e., few-shot or zero-shot) labelled examples~\cite{brown2020language, kojima2022large}. %Examples such as~\cite{hegselmann2023tabllm} demonstrate that LLMs consistently outperform supervised classifiers such as logistic regression and LightGBM in the classification of healthcare claims tabular datasets in the zero and few-shot settings.  \textcolor{red}{on tabular datasets? Based on the flow this work is better explained in the next paragraph}

%This has driven the development of multiple improved LLMs, such as the Megatron-turing NLG, Minerva and PaLM. \textcolor{red}{The last sentence is doesn't seem to add much value without details./Also these are a bit older models than GPTs etc.?} \textcolor{red}{What's more punchier would be to point out an example,where you highlight for a given example NLP task few-shot learning resulted in same levels of supervised learning.}

% Few shot classification therefore forms a sub-category under the in-context learning paradigm of LLMs' capabilities where a user instructs a black-box LLM for a classification task and provides a limited number of labeled samples as examples within the prompt \textcolor{red}{these sentences are kind of duplicating with what you already said in the first paragraph}. 

Various works applied the in-context learning capabilities to solve a range of problems, beyond typical natural language processing tasks. For instance, Hegselman et al.~\cite{hegselmann2023tabllm} and Jaitly et al.~\cite{jaitly2023towards} applied LLM-based few-shot classification to tabular data. The authors evaluate various serialisation methods and demonstrate the superior performance of few-shot LLMs in comparison to deep learning-based classification on tabular datasets. Other works such as~\cite{li2024cancergpt, yang2024uniaudio, van2023detecting} applied few-shot learning for domain-specific classification tasks. For instance, Li et al.~\cite{li2024cancergpt} evaluated the application of LLMs in biological inference. They prompted LLMs to predict the synergy of drug pairs in rare tissues by including a few labeled examples in the prompt. In contrast, Yang et al.~\cite{yang2024uniaudio} applied few-shot LLMs in a cross-modal framework for several audio tasks such as speech emotion classification, audio classification, and text-to-speech classification, while Van et al.~\cite{van2023detecting} investigated the few-shot capabilities of vision language models on hateful meme detection. 

%For example, Li et al.~\cite{li2024cancergpt} assessed the use of LLMs in biological inference by prompting LLMs to predict the synergy of drug pairs in rare tissues by including a few labelled examples in the prompt. 

%While LLM-based classification has been leveraged within the cyber security domain, it has not been applied to URL classification and existing works mainly focus on fine-tuning open LLMs rather that taking a zero-shot or few-shot approach. 

Several work utilised the capabilities of LLMs in cybersecurity. Nonetheless, the efforts have been much more focused on fine-tuning than few or zero-shot settings. For example, Aghaei and Al-Shaer~\cite{aghaei2023cve} proposed SecureBERT, which is a security-specific fine-tuned BERT model that achieves F1 scores ranging from 95\% to 98\% in classifying Common Vulnerabilities and Exposures (CVEs) to their corresponding MITRE ATT\&CK techniques. The authors also demonstrated that SecureBERT outperforms ChatGPT in Tactics, Techniques, and Procedures (TTP) prediction. Balasubramanian et al.~\cite{balasubramanian2023transformer} proposed finetuned GPT-3 models as anomaly detection classifiers which achieve more than 99\% accuracy. 

\textit{In contrast to these works, we propose a one-shot LLM classification approach for phishing URL detection, which can also be applied in zero-shot and few-shot settings. Our approach does not require fine-tuning of the language model parameters and demonstrates prediction performances comparable to supervised approaches.}

\subsection{Explainability of LLMs as classifiers}

The recent success of LLMs in varied domains has additionally motivated researchers to assess LLMs' capabilities as model explainers. Traditionally, explainability algorithms explain black-box model predictions to elucidate the inner model mechanisms. Metrics used to evaluate these algorithms, therefore, assess \textit{faithfulness} (i.e., how well the explanations reflect the true inner mechanism of the model) and \textit{plausibility} (i.e., how well the explanations correlate with human reasoning), aiming to enhance interpretability and trust in the otherwise black-box automated decision-making process. Studies that assess the explainability of LLMs include~\cite{kroeger2023large, bhattacharjee2024towards}, which implements LLMs as post-hoc explainers of predictions generated by other models. Meanwhile, LLM chain-of-thought reasoning has been considered as self-explanations for their predictions/classifications~\cite{chen2023models} and in comparison to traditional explanation methods, LLM natural language explanations have been noted for their high plausibility. That is, these explanations are consistent with human logic and reasoning~\cite{agarwal2024faithfulness, turpin2024language}. Nonetheless, due to the complexity of large language model architectures and the variability in its outputs, the application-specific correctness of LLM text generation remains largely unexplored.

%Although LLMs are not trained to express the systematic, model-specific reasons for its outputs~\cite{tanneru2024hardness}, studies such as ~\cite{agarwal2024faithfulness} demonstrate that LLM self-explanations are characterised by high plausibility. That is, they are consistent with human reasoning. plausibility evaluation studies of LLM self-explanations such as chain-of-thought reasoning, have found them to be lacking in faithfulness~\cite{turpin2024language, tanneru2024hardness} while overemphasising plausibility~\cite{agarwal2024faithfulness} %\textcolor{red}{talking about something it lacks undermines your work here. Rewrite it.}. This is expected given LLMs' training paradigm, which encourages the generation of human-like reasoning and coherent explanations. 

\textit{Therefore, in this paper, we propose a method to bridge this gap for URL classification self-explanations. As our one-shot classification framework outputs both explanation and prediction, we assess the correctness of the URL benign and phishing indicators identified in the self-explanations by comparing them to post-hoc explanations obtained through a supervised training setting.}

\section{Our Framework}
\label{Sec:Approach}

\begin{figure*}[t]
     \centering
     \includegraphics[width=0.9\textwidth]{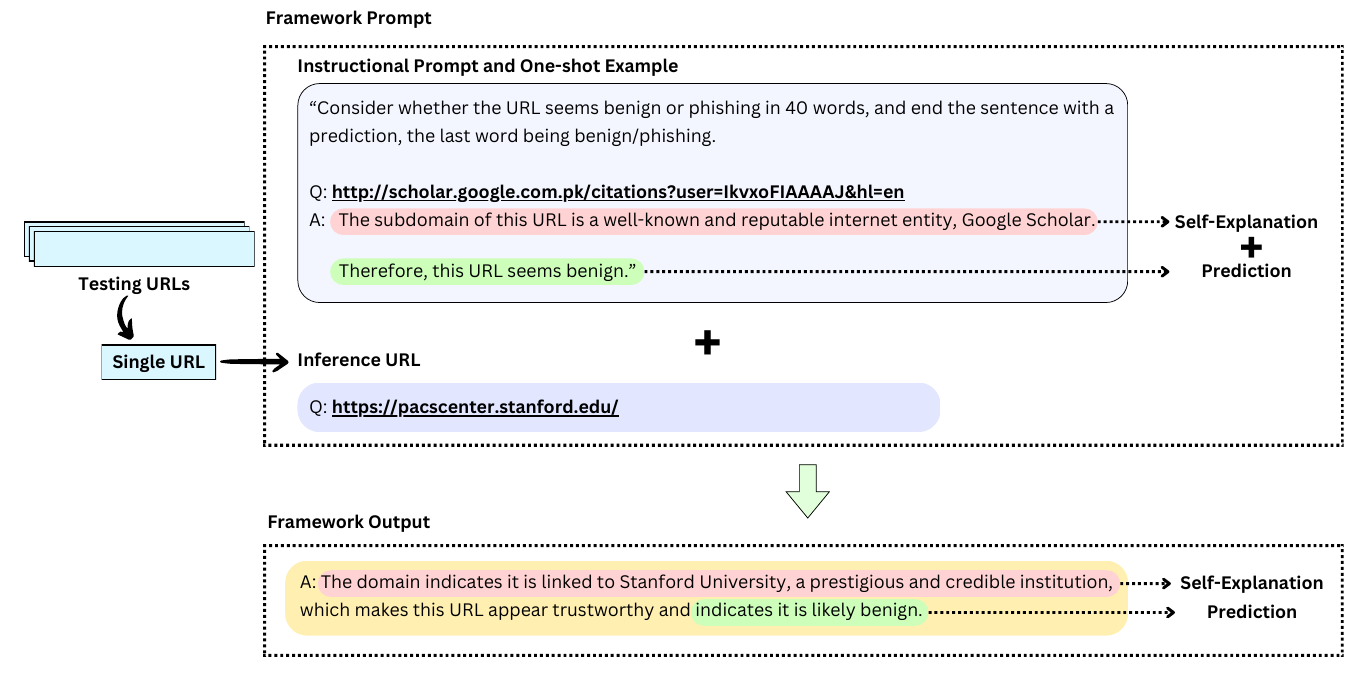}
     \caption{LLM-based One-Shot URL Classification Framework}
    \label{Fig:intro_framework}
\end{figure*}

As mentioned in the introduction, our aim is to address the problem of generalisability and explainability in phishing URL detection through a one-shot LLM-based classification framework. We implement a lightweight approach that leverages LLMs' vast embedded knowledge to improve classification accuracy while reducing false positive rates. This approach also provides explanations for each prediction to improve usability and empower end-users with more knowledge about a given URL prediction.

Our framework consists of prompting the LLM with the instruction and one example, as shown in Figure 1. Following the prompt we provide the URL for which we require an output from the LLM. The objective of this prompt is to not only obtain a single word prediction but also obtain the reasoning or explanation that leads to the prediction. In particular, we prompt a targeted form of LLM reasoning by specifically instructing the LLM to consider the benign and phishing characteristics of the URL. This form of targeted reasoning taps into LLM knowledge regarding legitimate entities and websites present on the web. It also encourages the LLM to consider the structure of the URL and whether it reflects phishing characteristics. 

In finalising our framework we faced the following technical challenges:

\begin{itemize}
  \item{\bf Prompting for classification:} To query an LLM about whether a given URL is benign or phishing we experimented with two variations of prompting. The first type gave the LLM the following four options to choose from when deciding a classification - benign/phishing/potentially phishing/uncertain. This was motivated by the idea that allowing more granular labelling options would drive the LLM to make more precise decisions. We processed the LLMs’ outputs to this prompt by considering all benign labelled URLs as benign and the rest as phishing. That is, it is safer to consider a URL as phishing if there is any uncertainty or potential for harm. The second type of prompting we tried is the one used in our proposed framework. Our initial analysis showed that the second prompt style returned slightly higher accuracy than the first prompt. The reason for this is that the availability of the ‘potentially phishing’ and ‘uncertain’ label options encouraged the LLM to take a cautious approach during classification. As a result, many benign URLs with less well-known domains were labelled as ‘uncertain’ or ‘potentially phishing’, thereby increasing the false positive rate. In contrast, the second prompt type (our proposed framework) encouraged the LLM to be more decisive about benign URL domains which improved the overall classification accuracy. 
  \item{\bf Chain-of-Thought reasoning:} Some of our initial experiments verified that using chain-of-thought (CoT) reasoning before making a prediction improved detection accuracy. To implement this in our prompt we experimented with two versions of CoT prompting. The first version prompted the LLM to consider the domain, subdomain and path of the URL separately and identify benign and phishing characteristics before making a prediction. The second version simply prompted the LLM to consider the benign and phishing characteristics (this is the version used in our final proposed method). With the first CoT prompt version, we found that explicitly identifying benign and phishing characteristics in the various URL components caused the LLM to place unnecessary weight on these features which resulted in lower overall prediction accuracy. Instead allowing the LLM more flexibility in its reasoning by simply prompting it to consider the general benign and phishing characteristics of the URL (second CoT prompt version) resulted in higher overall detection accuracy. 
  \item{\bf Number of examples to include in prompt:} We experimented with varying numbers of few-shot examples to assess if increasing the number of examples improved the prediction accuracy. Besides the one-shot approach, we experimented with two examples (one phishing URL and one benign URL) and five examples (three benign URLs and two phishing URLs). Our analysis showed that increasing the number of examples did not improve the prediction accuracy by a significant amount and therefore decided on a one-shot approach. 
\end{itemize}

Figure~\ref{Fig:intro_framework} illustrates the overall flow of our framework. In particular, we show how we consider the first part of the output to be a \textit{self-explanation} for the prediction that follows it. In the following section, we evaluate the performance of five large language models in returning accurate predictions for URLs in such a one-shot setting. We also evaluate the quality of the self-explanations that accompany the prediction. 

Here, note that we operate in a one-shot setting, i.e., we use only one URL (i.e., the Google Scholar URL as the sole example) to generate all of our results in Section~\ref{Sec:Results}. We emphasise that our prompt is the same for all of our results (i.e., it is the same query with the same Google Scholar URL). Later in Section~\ref{SubSec:FewShot}, we show the viability of our approach in zero-shot and few-shot settings as well.

\section{Experiment Settings}
\label{Sec:predacc_eval}

In this section, we first introduce the datasets we use to evaluate the prediction accuracy of our framework. Next, we describe how we evaluate the prediction performance of LLMs as one-shot URL classifiers in comparison to supervised URL classifiers, followed by the experiment settings we use to evaluate the quality of LLM self-explanations.

\subsection{Datasets}
\label{SubSec:Datasets}

We use two publicly available URL datasets and another URL dataset that was collected as part of our previous work~\cite{rashid2024phishing}. \\ \vspace{-3mm}

\noindent{\bf 1) ISCX-2016 Dataset:} Collected by Mamun et al.~\cite{mamun2016detecting}, this dataset consists of 35,300 benign URLs collected by crawling Alexa top domains and various malicious URL types, of which 9,964 are phishing URLs sourced from OpenPhish.\footnote{openphish.com} Although the dataset contains more than 100,000 URLs altogether, we randomly selected 10,000 benign URLs and combined them with the phishing URLs to obtain a more balanced dataset. \\ \vspace{-3mm}

\noindent{\bf 2) EBBU-2017 Dataset:} This dataset was introduced in~\cite{sahingoz2019machine}, comprising 36,400 benign URLs and 37,175 phishing URLs. The benign URLs were collected by querying a specific list of words, constructed by the authors of the paper, on the Yandex search engine\footnote{yandex.com} and selecting the highest-ranked URLs that were returned from this query. The phishing URLs were collected from PhishTank.\footnote{phishtank.org} \\ \vspace{-3mm}

\noindent{\bf 3) HISPAR-Phishstats (HP) Dataset:} This dataset was introduced in our previous work~\cite{rashid2024phishing}. We collected 46,250 benign URLs from the HISPAR list~\cite{aqeel2020landing} such that the set comprised 1,850 unique domains from the Alexa top domains list with 25 different URLs for each domain. The dataset also consisted of 46,250 unique phishing URLs, which we collected from Phishstats\footnote{phishstats.info} between the 19th of June and the 26th of August 2022. 

%\textcolor{blue}{For the ISCX-2016 and EBBU-2017 datasets which are older, the existence of well known benign URLs can introduce some bias in the predictions made by the LLMs which are likely to have seem the benign URLs during training. Unlike the ISCX-2016 and EBBU-2017 datasets, this HP dataset is relatively new as we assembled it recently in our previous work. It is therefore unlikely that the LLMs gained knowledge about this dataset during training.}

To train the supervised training baselines, we randomly split these datasets into training, validation, and test sets in a 60:20:20 ratio as summarised in Table~\ref{Tab:Datasets}. To evaluate our LLM-based framework, we use a randomly sampled subset of 1,000 URLs from the test splits of each dataset. We limit our analysis to 1,000 samples due to cost considerations when accessing some of the LLM APIs.

\begin{table}[!ht]
    \centering
    \scriptsize
    \caption{Dataset summary}
    \begin{tabularx}{\columnwidth}{X X X X X X X}
    %\begin{tabular}{|l|l|l|l|l|l|l|}
    \hline

        \multirow{2}{*}{\textbf{Dataset}} & \multicolumn{2}{c}{\textbf{Training}} & \multicolumn{2}{c}{\textbf{Validation}} & \multicolumn{2}{c}{\textbf{Test}}  \\%\cline{2-7}
        &  \multicolumn{1}{c}{\textbf{Benign}} & \multicolumn{1}{c}{\textbf{Phishing}} & \multicolumn{1}{c}{\textbf{Benign}} & \multicolumn{1}{c}{\textbf{Phishing}} & \multicolumn{1}{c}{\textbf{Benign}} & \multicolumn{1}{c}{\textbf{Phishing}} \\ \cline{2-7}%\hline
        ISCX & 5,978 & 6,001 & 2,043 & 1,950 & 1,979 & 2,014 \\ %\hline
        EBBU & 21,804 & 22,341 & 7,227 & 7,488 & 7,369 & 7,346 \\ %\hline
        HP & 27,740 & 27,760 & 9,180 & 9,320 & 9,330 & 9,170 \\ \hline
    %\end{tabular}
    \end{tabularx}
    \label{Tab:Datasets}
\end{table}

\subsection{Accuracy comparison with supervised URL classifiers}
\label{SubSec:AccuracyExperiment}

We evaluate the prediction performance of LLMs under our framework by comparing their accuracy to state-of-the-art URL classifiers trained in a fully supervised setting. For this, we refer to our previous work\cite{rashid2024phishing} where we trained four state-of-the-art URL classifiers: URLNet~\cite{le2018urlnet}, URLTran~\cite{maneriker2021urltran}, CatchPhish~\cite{rao2020catchphish} and PhishRF~\cite{sahingoz2019machine}. We trained and tested the performance of these classifiers using the training and testing datasets described in Table ~\ref{Tab:Datasets}. We present the performance of the supervised URL classifiers when applied to each test dataset in Table~\ref{Tab:baselines}. Here, note that the models work well when they are tested on the test set which comes from the same dataset. For example, the URLtran model trained on the ISCX dataset achieves an F1 score of 0.99 when tested on the ISCX test set. However, its performance drops significantly to F1 scores of 0.69 and 0.68 when tested on the EBBU and HP test sets, respectively. 

We obtain the one-shot prediction accuracies under our framework using the 1,000 randomly selected samples of the test datasets in Table ~\ref{Tab:Datasets} as described in Section~\ref{SubSec:Datasets}. That is, in each query, we prompted the LLMs to reason and predict the label of one given URL from the test sets. As our framework specifically instructed for the prediction to be stated in the final word of the output as either `benign' or `phishing' we considered each output prediction to be either benign or phishing if the last 20 characters of the output contained the terms `benign' and `phishing' respectively. If the last 20 characters contained neither of the terms we considered the prediction to be `uncertain'. However, in the calculation of accuracy we counted the `uncertain' predictions as phishing, considering that would be the safer decision in a practical setting.

We tested five state-of-the-art LLMs - GPT 4-Turbo~\cite{achiam2023gpt}, Claude 3 Opus~\cite{anthropic2024claude}, Gemini~\cite{team2023gemini}, LLaMA 3~\cite{llama3}, and LLaMA 2~\cite{touvron2023llama}. We repeated our experiments for each LLM and each test dataset five times and calculated the average F1 score. We select the F1 score as a performance metric as it accounts for the true positive rate as well as the false positive rate. We compare the F1 scores of each LLM and each test set under our framework with the F1 scores obtained by supervised URL classifiers in Table~\ref{Tab:baselines}. Since LLMs can generate different outputs to the same query at different times, for each of the 1,000 URLs, we queried the LLM five times and report the average performance results together with the standard deviation. We present these results in Section~\ref{SubSec:FewShot}.

%\textcolor{red}{Is there a reason why we include F1 score than the accuracy? I think it is because of the positive class. If there is an explanation, add it here.}

%Our main objective is to assess the few-shot predictive capability of LLMs on a test dataset in comparison to URL classifiers trained in different supervised settings. That is, we compare the cross-dataset test accuracy of supervised URL classifiers trained on three different datasets with the prediction accuracy of LLMs using our framework. Here, we refer to our previous work, demonstrating the cross-dataset performance state-of-the-art URL classifiers (Table 1){cite}. We demonstrated that while the URL classifiers return high F1 scores when the test set and training set are from the same source, the prediction performance drops by 10-30\% when the test URLs are from a source that is different to that of the training URLs. Our analysis showed that the cause of this that the URL classifiers are not robust to domain shifts amongst URL datasets procured from different sources. This motivates our few-shot approach to use LLMs as URL classifiers without the need of a large training dataset. 

\begin{table}[t]
\scriptsize
    \centering
    \caption{F1 scores for baseline experiments using the URLNet, URLTran, CatchPhish, and PhishRF models.}
    %\begin{tabularx}{\columnwidth}{|l|l|l|l|l|}
    \begin{tabular}{l l l l l}
    %\hline
        ~ & ~ & \multicolumn{3}{c}{\textbf{Test Sets}} \\ %\hline
        %~ & ~ & Test Sets   &  & \\ \hline
        \textbf{Train Sets} & \textbf{Models} & ISCX & EBBU & HP \\ \hline
        \multirow{4}{*}{ISCX} & URLNet & \textbf{0.99} & 0.69 & 0.68 \\ %\cline{2-5}%\hline
             & URLTran & \textbf{0.99} & 0.68 & 0.67 \\ %\cline{2-5}%\hline
             & CatchPhish & \textbf{0.99} & 0.68 & 0.69 \\ %\cline{2-5}%\hline
             & PhishRF & \textbf{1.00} & 0.68 & 0.67 \\ \hline
        \multirow{4}{*}{EBBU} & URLNet & 0.88 & \textbf{0.98} & 0.92 \\ %\cline{2-5}%\hline
             & URLTran & 0.83 & \textbf{0.99} & 0.93 \\ %\cline{2-5}%\hline
             & CatchPhish & 0.80 & \textbf{0.94} & 0.88 \\ %\cline{2-5}%\hline
             & PhishRF & 0.87 & \textbf{0.96} & 0.94 \\ \hline
        \multirow{4}{*}{HP} & URLNet & 0.72 & 0.79 & \textbf{0.99} \\ %\cline{2-5}%\hline
           & URLTran & 0.75 & 0.86 & \textbf{0.99} \\ %\cline{2-5}%\hline
           & CatchPhish & 0.73 & 0.72 & \textbf{0.98} \\ %\cline{2-5}%\hline
           & PhishRF & 0.83 & 0.86 & \textbf{0.99} \\ \hline
    \end{tabular}
    \label{Tab:baselines}
    %\end{tabularx}
\end{table}

\subsection{Quality evaluation of LLM Self-explanations}
\label{eval_explanations}

We evaluate the quality of the LLM self-explanations from two aspects.
First, we compare the benign and phishing indicators given by LLMs with the indicators returned by a post-hoc explanation method (i.e., LIME~\cite{ribeiro2016should}) applied to the corresponding supervised URL classifier. This is to justify further that the few-shot predictions the LLMs make are logical and often have the same reasoning as the corresponding supervised learning classifier that has trained on large volumes of training data. Second, we evaluate the general quality of the LLM-generated explanations in terms of readability, coherence, and informativeness using the G-Eval~\cite{liu2023g} framework. %We present the results in Section~\ref{results}.

%For this analysis, we use \texttt{GPT-4o}, which was the best-performing model in the results we presented in Section~\ref{results}}.

\subsubsection{Alignment between LLM and LIME indicators}

Classifying a given URL as benign or phishing is an overall decision made by the classifier observing various indicators of a given URL. The datasets in this area ({\bf cf.} Table~\ref{Tab:Datasets}) consist of only URLs and their corresponding labels. Therefore, there is no gold-standard ground truth to the explanations provided by the LLM on benign and phishing indicators.

As a result, we establish a proxy metric by comparing the LLMs' self-explanations to the indicators returned by a post-hoc explanation method applied to a supervised URL classifier. Specifically, we select URLTran as it returned the highest average F1 score across the three datasets ({\bf cf.} Table~\ref{Tab:baselines}) in the supervised setting. We applied the LIME algorithm~\cite{zhang2019should} to three separate URLTran models trained on each of the three training sets and obtained LIME indicators for the models' predictions for their respective test sets.

The LIME algorithm determines feature attributions of a given test sample by approximating a sparse linear model based on the model predictions for inputs sampled from the neighbourhood of that test sample. Feature importance is then determined by the feature coefficients of that linear model. As each URLTran model returned F1 scores of 0.99 for their respective test sets, we assume that the sparse linear model learned by LIME and, therefore, the indicators it returned are a true reflection of the benign and phishing characteristics represented within labelled URL datasets. 
%\textcolor{red}{need a definition and probably a citation for this} 

By comparing the LLM self-explanations with the LIME indicators, we can verify how well benign and phishing characteristics identified by LLMs align with label-specific features learnt by supervised models from URL-labelled datasets. We measure this alignment using Jaccard similarity. 

%Below we provide an example of a given URL, its corresponding explanation and the self-explanation indicators obtained using the prompt in Figure ~\ref{llm_indicator_prompt}. \\ \vspace{-3mm}

%GPT-4o (i.e., the best-performing model out of the LLMs we tested - \textcolor{red}{is this correct? Did we use GPT4 or GPT-4o? Bacuse later you say GPT-4o is outside our testing ones?})

  %\textcolor{red}{expand this using an example.Write things like a high Jaccard similarity will indicate an overlap between the LLM predictions etc.}

\begin{tcolorbox}[colback=gray!10!white, colframe=black, title=LLM self-explanation Example]

``You are an intelligent assistant. For the given URL, return only the specific URL parts that the explanation considers benign or phishing. You must use the format `Benign:Phishing:". \\ \vspace{-3mm}

\small{}
{\bf Q. URL:} \\ \vspace{-4mm}
%\url{http://www.vivscreisveci.vcirveseiaveesi.ghqphy.top/uWBRvZ8quj/page1.php} \\

\fontsize{8pt}{10pt}\selectfont % Adjust the font size here
\texttt{\url{http://www.vivscreisveci.vcirveseiaveesi.ghqphy.top/uWBRvZ8quj/page1.php}} \\ \vspace{-3mm}

\small{}
%\texttt{\url{http://www.vivscreisveci.vcirveseiaveesi.ghqphy.top/uWBRvZ8quj/page1.php}} \\

{\bf Exp:} The URL contains a suspect domain that is not
recognizable and appears to have randomly generated parts,
which is typical of phishing sites aiming to mimic legitimate
ones deceptively. This URL seems to be phishing. \\ \vspace{-3mm}

{\bf A. Benign:} \url{`'} \small {\bf Phishing:}  \fontsize{8pt}{10pt}\selectfont \fontsize{8pt}{10pt}\selectfont \texttt{\url{`vivscreisveci.vcirveseiaveesi.ghqphy.top'}},\texttt{\url{`uWBRvZ8quj'}} \\ \vspace{-3mm}

\small{}

{\bf Q. URL:} \\
\fontsize{8pt}{10pt}\selectfont \texttt{\url{https://reciclatex.com/ES/cx/home}}\\ \vspace{-3mm}

\small{}

{\bf Exp:} This URL seems suspicious because it lacks recognizable
branding and could be mimicking a legitimate site with its
structure, but without contextual evidence of malicious intent,
it's uncertain. My prediction leans towards benign. \\ \vspace{-3mm}

{\bf A. Benign:} \fontsize{8pt}{10pt}\selectfont  \texttt{\url{`com', `ES',`https'}} \small {\bf Phishing:} \fontsize{8pt}{10pt}\selectfont 
 \texttt{\url{`reciclatex', `home', `cx'}} \\ \vspace{-3mm}
%\normalsize{}

%Benign: 'com', 'ES', 'https' Phishing: 'reciclatex', 'home', 'cx' "

\end{tcolorbox}
\noindent\begin{minipage}{.5\textwidth}
\captionof{figure}{Prompting the LLM to list benign and phishing indicators identified in the self-explanation}\label{llm_indicator_prompt}
\end{minipage}

Specifically, we queried each LLM with the prompt in Figure~\ref{llm_indicator_prompt}. We followed this prompt with the URLs from our test set and their corresponding self-explanations from the respective LLMs, one at a time, to obtain the self-explanation indicators. For each URL, we compared the self-explanation indicators with the LIME indicators that were returned for the same URL from the corresponding URLTran model. For example, if the given URL belonged to the HP test set, we compared the LLM self-explanation indicators to the LIME indicators for the same URL when applied to URLTran trained on HP training set. \\ \vspace{-3mm}

\noindent{\textbf{Jaccard Similarity}: To numerically compare the two sets of indicators (i.e., LLMs and LIME), we used the Jaccard similarity coefficient~\cite{jain1988algorithms}, which calculates a similarity score between two sets as the ratio of the intersection of the sets and their union: $J(A,B) = \frac{|A \cap B|}{|A \cup B|}$. Here, the label-specific LIME indicators form one set of elements and the indicators identified by the LLM self-explanation form another set. As each self-explanation was in natural language, we extracted the specific URL components that it identified using LLM prompting.}

\noindent{\textbf{Example Jaccard Similarity Calculation}: 

\begin{itemize}
    \item URL: \textit{\url{https://reciclatex.com/ES/cx/home}} 
    \item Explanation: ``This URL seems suspicious because it lacks recognizable branding and could be mimicking a legitimate site with its structure, but without contextual evidence of malicious intent, it's uncertain. My prediction leans towards benign."
    \item Self-explanation indicators: \textit{Benign: `com', `ES', `https' Phishing: `reciclatex', `home', `cx'}
\end{itemize}

As the original LLM output predicted the URL as benign, we collect the benign self-explanation indicators delimited by special characters (A: \textit{`com', `ES', `https'}) and compare them with corresponding benign LIME indicators (B: \textit{`com', `ES', `reciclatex', `https', `home'}). Using the Jaccard similarity measure, this corresponds to:

% \footnotesize
% \begin{equation}
% \begin{split}
% J(A,B) & =\frac{|\{'com', 'ES', 'https'\}|}{|\{'com', 'ES', 'https', 'home', 'reciclatex'\}|} \\
%  & = 0.6
% \end{split}
% \end{equation}
% \normalsize{}

\small
\begin{equation}
J(A,B) = \frac{|\{ \text{\textit{`com'}}, \text{\textit{`ES'}}, \text{\textit{`https}} \}|}{|\{ \text{\textit{`com'}}, \text{\textit{`ES'}}, \text{\textit{`https'}}, \text{\textit{`home'}}, \text{\textit{`reciclatex'}} \}|} = 0.6
\end{equation}
\normalsize{}

The Jaccard similarity score ranges from 0 to 1, where higher similarity scores indicate a greater overlap between self-explanation indicators and LIME indicators. We conduct the above evaluation for each of the LLMs that we evaluate under our framework and for each of the three datasets. We report these results in Section~\ref{alignment_results}.
  
%\textcolor{red}{Can you make this example to build on what is shown in Figure 3? Otherwise, it is too much repetition that we have another example for this calculation.}

\begin{figure*}[t]
     \centering
     \includegraphics[width=0.7\textwidth]{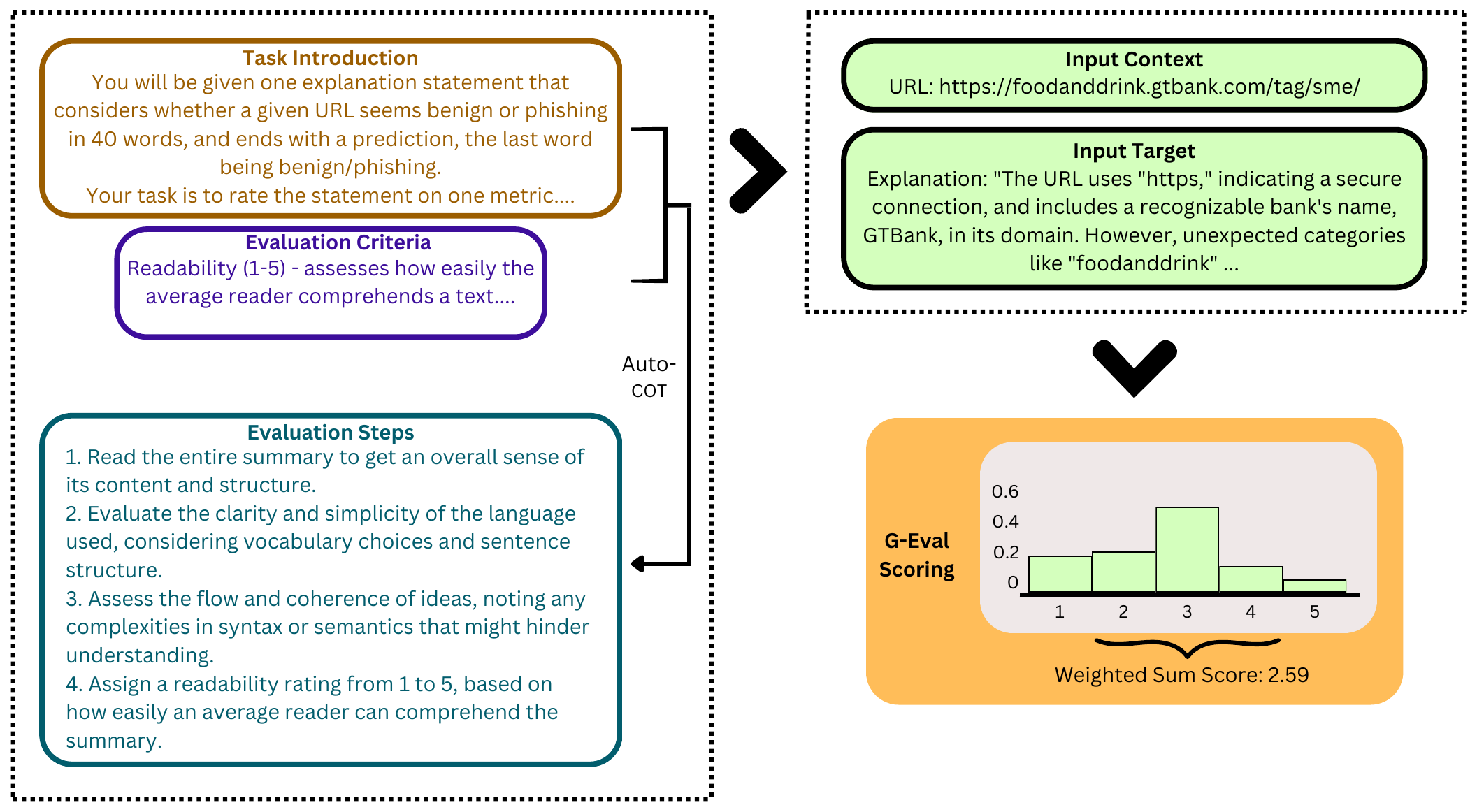}
     \caption{Using the G-Eval Framework to assess the quality of LLM self-explainations}
    \label{geval_framework}
\end{figure*}

\subsubsection{G-Eval Framework}
To evaluate the general quality of LLM outputs, we use the G-Eval framework~\cite{liu2023g}, which implements LLMs to evaluate the quality of a text. Specifically, the framework uses a chain-of-thought approach to use an LLM to evaluate a text against a given metric. In order to be useful to the end-users of a URL classification system, we assess the self-explanations according to the following metrics: \textit{readability}, \textit{coherence}, and \textit{informativeness}. We adopt the definition of `readability' provided in~\cite{agrawal2024cyberq} for cyber security question-answering tasks, and we adapt the definitions of `coherence' and `informativeness' from~\cite{celikyilmaz2020evaluation} to suit our application as described below.

  \begin{itemize}
  \item {\bf Readability (1-5):} Assesses how easily the average reader comprehends a text, considering factors like lexical, syntactic, semantic, and stylistic complexity. 
  \item {\bf Coherence (1-5):} Assesses the collective quality of all sentences. This metric measures how well the statement is structured and organised in explaining why the URL is predicted to be either benign or phishing.
  \item {\bf Informativeness (1-5):} Measures how well the output answers the question. That is, this metric assesses how clearly the statement considers the benign and phishing characteristics of the URL and finally provides a prediction.
  \end{itemize}

In the G-Eval framework, an LLM evaluates a given text against a given criteria and the criteria definition by following a sequence of evaluation steps generated by the LLM itself through chain-of-thought reasoning.  That is, an LLM is given a criterion and its definition and prompted to produce a series of steps that it should follow to evaluate a text against that criterion for a specific task. As we are investigating five LLMs for prediction accuracy and explanation quality under our one-shot framework, we use a different LLM  (GPT-4o)  in our G-Eval implementation to evaluate the other five LLMs' self-explanations. 

Figure~\ref{geval_framework} which is adapted from the original paper~\cite{liu2023g} illustrates our use of G-Eval for the readability metric. We give the \textit{task introduction} and the \textit{evaluation criteria} to GPT-4o to generate evaluation steps. Then we input each test URL (given as context) and the corresponding self-explanation (the text to be evaluated), and the G-Eval framework calculates a probability-weighted summation of all possible scores (1-5). We followed a similar approach for the coherency and informativeness metrics. %We show the full prompt, the evaluation criteria, and the GPT-4o-generated evaluation steps for readability in Appendix A.

%For example, to evaluate the readability of the self-explanations under our one-shot framework, we gave the prompt in Figure~\ref{geval_steps_prompt} to GPT-4o to generate evaluation steps. The evaluation steps generated for readability, coherence and informativeness are provided in Appendix~\ref{App:eval_steps}.

%\textcolor{red}{Figure 5 is not referred in text?} %\textcolor{red}{don't you need to metion that you used two LLMs? In general you need to explain a bit more on the workings of G-Eval framework. It happens to some extent in the below example, but it is good enough for an outsider to understnad it. You night need a flow chart/block diagram for it, with showing these text next to the corresponding block.}

%Figure~\ref{geval_framework} is adapted from the original paper ~\cite{liu2023g} to reflect our implementation of G-Eval for our few-shot URL classification framework. It shows the evaluation steps generated for the prompt in Figure~\ref{geval_steps_prompt} and how these are then used by GPT 4o to evaluate the self-explanation of a given URL.  

We generate readability, coherence, and informativeness scores for the self-explanations corresponding to all URLs from the three test sets generated by the five LLMs that we evaluate under our one-shot framework. We report these results in Section~\ref{geval_results}.

\section{Results}
\label{Sec:Results}

\subsection{Prediction performance of one-shot LLM URL classifiers}
\label{SubSec:FewShot}

\begin{table*}[t]
    \centering
    \small
    \caption{Prediction performance of one-shot classification - F1 Scores}
    \begin{tabularx}{\columnwidth}{X X X X X X X}
    %\begin{tabular}{|l|l|l|l|l|l|l|}
    \hline

        \multirow{2}{*}{\textbf{Test Dataset}} & \multicolumn{6}{c}{\textbf{Models}} \\%& \multicolumn{2}{c}{\textbf{Validation}} & \multicolumn{2}{c}{\textbf{Test}}  \\%\cline{2-7}
        &  \multicolumn{1}{c}{\textbf{URLTran (Supervised)}} & \multicolumn{1}{c}{\textbf{GPT 4}} & \multicolumn{1}{c}{\textbf{Claude 3}} & \multicolumn{1}{c}{\textbf{Gemini}} & \multicolumn{1}{c}{\textbf{LLaMA 3}} & \multicolumn{1}{c}{\textbf{LLaMA 2}} \\ \cline{2-7}%\hline
        HP & 0.99 & $\textbf{0.94} \pm 0.00153$ & $0.89 \pm 0.00092$ & $0.87\pm0.00529$ & $0.84\pm0.00358$ & $0.67\pm0.00215$ \\ %\hline
        EBBU & 0.99 & $\textbf{0.95}\pm0.00393$ & $0.91\pm0.00184$ & $0.86\pm0.01191$ & $0.87\pm0.00378$ & $0.69\pm0.00220$ \\ %\hline
        ISCX & 0.99 & $\textbf{0.87}\pm0.00471$ & $0.85\pm0.00284$ & $0.78\pm0.00589$ & $0.81\pm0.00348$ & $0.67\pm0.00202$ \\ \hline
    %\end{tabular}
    \end{tabularx}
    \label{Tab:oneshot_acc}
\end{table*}

In Table ~\ref{Tab:oneshot_acc}, we present the F1 scores of our one-shot URL classification framework on the five LLMs we test ({\bf cf.} Section~\ref{SubSec:AccuracyExperiment}): GPT 4-Turbo, Claude 3 Opus, Gemini, LLaMA 3, and LLaMA 2. As explained in Section~\ref{SubSec:AccuracyExperiment}, these results are the average F1 score calculated over five repeats of the experiment.  For ease of comparison, we also present the F1 scores obtained by URLTran (i.e., best-performing model under the supervised setting - {\bf cf.} Table~\ref{Tab:baselines}) when applied to each test set after being trained on the respective training set. 

%of the following five state-of-the-art large language models under our one-shot URL classification framework: GPT 4-Turbo, Claude 3 Opus, Gemini, LLaMA 3 and LLaMA 2. Table ~\ref{Tab:oneshot_acc} presents the F1 scores obtained by the LLMs for the URLs from each test dataset (ref. Table ~\ref{Tab:Datasets}), averaged over five repeats, and the standard deviations. As URLTran was the best-performing model under the supervised setting (ref. Table ~\ref{Tab:baselines}), for ease of comparison we also present the F1 scores obtained by URLTran when applied to each test set after being trained on the respective training set. 

We observe that among the LLMs, GPT 4-Turbo produces the highest average F1 scores across all three datasets, with an overall average of 0.92. This is only 0.07 less than the performance of the fully supervised learned classifier. Claude demonstrates the next best performance with an overall average of 0.88, while Gemini and LLaMA 3 perform similarly, with overall averages of 0.83 and 0.84, respectively. Meanwhile, LLaMA 2 returns the lowest overall average F1 score of 0.68. The standard deviations over five repeats are very low for all the LLMs portraying overall consistency in accuracy over the five independent runs. 

%We provide an example of this in Section~\ref{geval_results}.

A more detailed analysis of the LLM outputs revealed that one of the reasons for LLaMA 2's significantly low performance may be its inability to generate good self-explanations. For many of the URLs, LLaMA 2 returned a prediction without considering the benign and phishing characteristics of the URL, as our prompt had asked.  Furthermore, even when the LLMs did consider the benign and phishing characteristics of the URL, in comparison to the other LLMs, GPT 4-Turbo returned more accurate predictions. Consider the following LLM outputs for the benign URL \textit{\url{https://reconciliation.americanexpress.com/}}.

\begin{itemize}
    \item {\bf GPT 4-Turbo:} \textit{``The domain hints at a well-known financial institution, American Express, but the use of "reconciliation" might raise suspicions without further context. Despite potential concerns, the domain's legitimacy suggests the URL is likely benign.''}
    \item {\bf Claude 3:} \textit{``The URL appears to be a legitimate American Express subdomain for account reconciliation. However, without further context, it's best to be cautious. Visit the main American Express website directly to avoid potential phishing attempts. The URL seems benign, but caution is advised.''}
    \item {\bf Gemini:} \textit{``The subdomain of this URL is reconciliation which is not a common subdomain of the American Express website. Therefore, this URL seems phishing.''}
    \item {\bf LLaMA 3:} \textit{``The URL appears to be a legitimate domain from a well-known financial institution, American Express. However, the specific page seems to be a secure login page, which may raise some concerns.''}
    \item {\bf LLaMA 2:} \textit{``The subdomain of this URL is American Express, which has been associated with phishing attacks in the past. Therefore, this URL seems phishing.''}
\end{itemize}

From the above examples, we observe that all LLMs are able to correctly identify the main entity in the URL domain as well as the context of the entity (financial institution). However, knowledge of the fact that many phishing attacks are financial attacks leads LLaMA 2 to directly classify the URL as phishing, which is an incorrect prediction. Meanwhile, Gemini incorrectly decides that the subdomain is not a common or expected one for the American Express website, which also leads it to predict the URL as phishing. While Claude 3 and LLaMA 3 correctly acknowledge that the domain and subdomain are legitimate, similar to LLaMA 2 and Gemini, the risk posed by the financial context of the URL, as well as the connection to account login/reconciliation, prevents the LLMs from classifying the URL as benign. In contrast to all the above, GPT 4-Turbo acknowledges the potential for suspicion given both of the above factors but proceeds to predict the URL as benign with apparent confidence (``likely benign").

\begin{figure*}[t]
     \centering     \includegraphics[width=\linewidth]%{images/hp_jacc_sim.pdf}
     {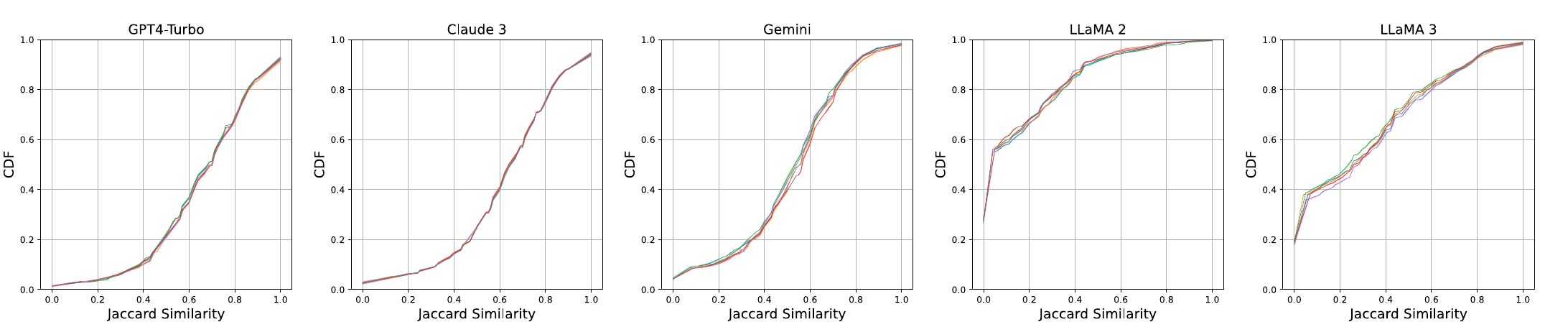}
     \caption{Cumulative distribution of the Jaccard similarity between LIME and LLM indicators - HP Dataset}
    \label{jacc-sim-cdf}
\end{figure*}

Another example demonstrates that LLaMA 3 and LLaMA 2 do not follow the correct reasoning based on their knowledge of legitimate entities, and this leads to an incorrect prediction. The benign URL \textit{\url{https://drfone.wondershare.net/ad/}} returned the following outputs from each LLM:

\begin{itemize}
    \item {\bf GPT 4-Turbo:} \textit{``The URL directs to a specific page of Wondershare's dr.fone, a recognized software for phone data management, implying it's a promotional link from a legitimate company. Consequently, this URL appears benign."}
    \item {\bf Claude 3:} \textit{``The URL is for a software product called Dr.Fone by Wondershare, which appears to be a legitimate data recovery tool. The .net top-level domain and lack of suspicious elements suggest this is likely a benign URL for the product's website or an advertisement. Benign."}
    \item {\bf Gemini:} \textit{``This URL belongs to a software company and therefore seems benign."}
    \item {\bf LLaMA 3:} \textit{``The URL appears to be a promotional link from a software company, Wondershare, which may be trying to entice users to download their product. Therefore, this URL seems phishing."}
    \item {\bf LLaMA 2:} \textit{``A: The subdomain of this URL is not a well-known or reputable internet entity, which suggests that it may be a phishing website. Prediction: Phishing."}
\end{itemize}

Again, we observe that GPT 4-Turbo recognises the domain and subdomain as legitimate and, therefore, classifies the URL as benign. Claude 3 and Gemini outputs follow a similar pattern. LLaMA 2, however, does not correctly recognise the \textit{Wondershare} entity, while for LLaMA 3, the possibility of the URL being a `promotional link' leads it to predict the URL as phishing.

Overall, the results show that the LLMs' self-explanation capabilities have an effect on the accuracy of their predictions. Specific to the phishing domains, these reasoning capabilities involve correctly identifying legitimate entities in the domains and subdomains of the URLs, as well as correctly attributing phishing and benign characteristics to the URLs. For example, despite correctly stating American Express has been associated with phishing attacks (as phishing URLs often include well-known entities in their URLs), LLaMA 2 failed to correctly identify the legitimacy of the specific American Express URL in the example above. %\textcolor{red}%{Is it also possible to comment that LLaMAs are trained on much smaller datasets than GPTs and others? with citations to justify.}

Finally, we highlight the low standard deviation values of the LLM F1 scores in Table~\ref{Tab:oneshot_acc}. This shows that LLMs are pretty consistent with their predictions over the five experiment runs, something we further analyse in Section~\ref{Sec:Consitency}.

\subsection{Quality of LLM outputs}

\subsubsection{Alignment between LLM self-explanations and LIME indicators - Jaccard Similarity}
\label{alignment_results}

As mentioned before, the purpose of this evaluation is to justify further that the one-shot predictions the LLMs make are logical and often have the same reasoning as the corresponding supervised learning classifier that has trained on large volumes of training data. From our repeated experiments to assess the prediction performance of our one-shot classification framework, we obtained five outputs for each URL from each of the five LLMs. Therefore, to evaluate the self-explanations produced by our framework, for each URL, we applied our Jaccard similarity evaluation to each of the five repeats for each URL. In Figure~\ref{jacc-sim-cdf}, we present the cumulative distributive functions (CDF) of the Jaccard similarity scores obtained for each LLM applied to the HP dataset. Each curve in each figure represents the CDF of the Jaccard Similarity values obtained from a single run. For brevity, we present results only for the HP dataset, and the other two datasets had similar patterns as the HP dataset. 
% ({\bf cf.} Appendix B)
We observe that the CDFs for GPT 4-Turbo and Claude 3 follow a strong sigmoid trend for each URL dataset, indicating they are distributed more towards high values (i.e., high alignment between LIME and LLM indicators). 

%s under our framework. Each line represents a distinct run/repeat of the experiment. 

More concretely, for each URL, we calculate the mean Jaccard similarity score to reflect the overall alignment between its self-explanation indicators and LIME indicators. In Table~\ref{Tab:jacc-frac} we show the proportion of the Jaccard scores that were above 0.5 for each dataset and each model. Here, we find that for GPT 4-Turbo, 72\%, 69\%, and 56\% of the Jaccard similarity scores were above 0.5 for the HP, EBBU, and ISCX test sets, respectively. These proportions gradually decrease for Claude 3, Gemini and LLaMA 3. For LLaMA 2 we find the lowest proportions, reflecting that very few of its LLM self-explanations were aligned with the LIME indicators.

\begin{table}[!ht]
    \centering
    \small
    \caption{Percentage of Jaccard similarity scores $\geq$ 0.5}
    \begin{tabularx}{\columnwidth}{X X X X X X}
    %\begin{tabular}{|l|l|l|l|l|l|l|}
    \hline

        \multirow{2}{*}{\textbf{Data}} & \multicolumn{5}{c}{\textbf{LLM}}   \\%\cline{2-7}
        &  \multicolumn{1}{c}{GPT 4} & \multicolumn{1}{c}{Claude 3} & \multicolumn{1}{c}{Gemini} & \multicolumn{1}{c}{LLaMA3} & \multicolumn{1}{c}{LLaMA2}  \\ \cline{2-6}%\hline
        HP & \multicolumn{1}{c}{72\%} & \multicolumn{1}{c}{69\%} & \multicolumn{1}{c}{51\%} & \multicolumn{1}{c}{22\%} & \multicolumn{1}{c}{6\%}  \\ %\hline
        EBBU & \multicolumn{1}{c}{69\%} & \multicolumn{1}{c}{65\%} & \multicolumn{1}{c}{32\%} & \multicolumn{1}{c}{29\%} & \multicolumn{1}{c}{8\%}  \\ %\hline
        ISCX & \multicolumn{1}{c}{56\%} & \multicolumn{1}{c}{48\%} & \multicolumn{1}{c}{23\%} & \multicolumn{1}{c}{19\%} & \multicolumn{1}{c}{7\%} \\ \hline
    %\end{tabular}
    \end{tabularx}
    \label{Tab:jacc-frac}
\end{table}

% \begin{table}[!ht]
%     \centering
%     \small
%     \caption{Percentage of Jaccard similarity scores $\geq$ 0.5}
%     \begin{tabularx}{\columnwidth}{X X X X X X}
%     %\begin{tabular}{|l|l|l|l|l|l|l|}
%     \hline

%         \multirow{2}{*}{\textbf{Data}} & \multicolumn{5}{c}{\textbf{LLM}}   \\%\cline{2-7}
%         &  \multicolumn{1}{c}{\textbf{GPT 4}} & \multicolumn{1}{c}{\textbf{Claude 3}} & \multicolumn{1}{c}{\textbf{Gemini}} & \multicolumn{1}{c}{\textbf{LLaMA3}} & \multicolumn{1}{c}{\textbf{LLaMA2}}  \\ \cline{2-6}%\hline
%         HP & \multicolumn{1}{c}{72\%} & \multicolumn{1}{c}{69\%} & \multicolumn{1}{c}{51\%} & \multicolumn{1}{c}{22\%} & \multicolumn{1}{c}{6\%}  \\ %\hline
%         EBBU & \multicolumn{1}{c}{69\%} & \multicolumn{1}{c}{65\%} & \multicolumn{1}{c}{32\%} & \multicolumn{1}{c}{29\%} & \multicolumn{1}{c}{8\%}  \\ %\hline
%         ISCX & \multicolumn{1}{c}{56\%} & \multicolumn{1}{c}{48\%} & \multicolumn{1}{c}{23\%} & \multicolumn{1}{c}{19\%} & \multicolumn{1}{c}{7\%} \\ \hline
%     %\end{tabular}
%     \end{tabularx}
%     \label{Tab:jacc-frac}
% \end{table}

%For instance, for GPT4-Turbo, approximately 65\% of the samples had more than 0.5 Jaccard Similarity, and approximately 20\% of the samples had more than 0.8 Jaccard Similarity. Calude 3 also shows similar behaviour. This is not the case for Gemini, LLaMA 3 and LLaMA 2. For instance, only around 50\% and 10\% samples had Jaccard Similarity values higher than 0.5 and 0.8, respectively, for Gemini.  LLaMA 3 and LLaMA 2 numbers were much lower.

To further exemplify this, in Table~\ref{Tab:alignment_example} we show a case where for a given URL, GPT 4-Turbo and LLaMA~2 return contradicting predictions and self-explanations. For each self-explanation, we used the prompt in Figure~\ref{llm_indicator_prompt} to extract the self-explanation indicators. We then compared these indicators to the LIME indicators corresponding to the LLM prediction. Therefore, we compare the GPT self-explanation indicators with the LIME benign indicators and the LLaMA 2 self-explanation indicators with the LIME phishing indicators. We observe that regardless of the predictions returned by each LLM, GPT 4-Turbo identifies the label-specific indicators more accurately than LLaMA 2, which is expressed by the Jaccard similarity between the respective self-explanation indicators and the LIME indicators.

\textit{Overall, these results indicate that the self-explanations returned by GPT 4-Turbo and Claude 3 are more consistent with the benign and phishing characteristics represented in URL datasets. As observed in the examples shown in the previous section, this is likely due to GPT 4-Turbo and Claude 3's ability to describe specific URL components as benign and phishing indicators when reasoning the predictions that both make.} 

%In contrast, the CDFs for the other LLMs show a steep rise for lower Jaccard similarity values. This indicates a higher chance of obtaining low Jaccard similarity scores when comparing Gemini, LLaMA 3 and LLaMA 2 self-explanations to LIME indicators. 

%Therefore, for GPT 4-Turbo and Claude, the distribution of Jaccard similarity scores favours higher values. As the Jaccard similarity measure ranges from 0 to 1 with 1 representing a perfect match, higher Jaccard scores reflect a higher probability of alignment between LLM self-explanations and LIME indicators.

\begin{table*}[t]
\centering
\small
\caption{Examples of alignment between LLM self-explanation indicators and LIME indicators}
\begin{tabular}{p{1.5cm}|p{5cm}|p{3.5cm}|p{3cm}|p{1cm}}

\multicolumn{5}{l}{\textbf{URL: https://drfone.wondershare.net/ad/ Original label: Benign}}\\ \hline
\textbf{LLM} & \textbf{Self-Explanation} & \textbf{Self-Explanation \newline Indicators} & \textbf{LIME indicators} & \textbf{Jaccard Score}\\ \hline
GPT \newline 4-Turbo & \textit{``The URL directs to a specific page of Wondershare's dr.fone, a recognized software for phone data management, implying it's a promotional link from a legitimate company. Consequently, this URL appears benign.}'' & {\textit{`drfone.wondershare.net'}, \textit{`ad'}, \textit{`https'}} & {\textit{`wondershare'}, \textit{`net'}, \textit{`https'}, \textit{`ad'}}  & 0.8        \\ \hline

LLaMA 2 & \textit{``The subdomain of this URL is not a well-known or reputable internet entity, which suggests that it may be a phishing website. Prediction: Phishing.''} & {\textit{`drfone'}, \textit{`wondershare'}, \textit{`ad'}} & {\textit{`drfone'}} & 0.3 \\\hline

\end{tabular}
\label{Tab:alignment_example}
\end{table*}

% \subsubsection{Alignment between LLM outputs and LIME indicators - User Study}

% This section will portray the results obtained from our user study to evaluate the alignment between LLM outputs and LIME indicators. The survey presented URLs and their corresponding LLM self-explanations and LIME indicators and asked them to select from the multiple-choice options the answer which reflected the alignment between the two. As the multiple-choice options represented a Likert scale, we converted the previously categorical options to a numerical scale of 1 to 6 such that 1 represented 'Completely Unaligned' and 6 represented 'Completely Aligned'. In Figure~\ref{user_study_histograms} we display the distribution of scores for benign and phishing URLs separately. 

% \begin{figure}
%      %\centering
%      \begin{subfigure}[b]{\linewidth}
%          \centering
%          \includegraphics[width=\linewidth]{images/benign_histogram.png}
%          \caption{Benign URLs}
%          %\label{fig:phishrf_feat_distr}
%      \end{subfigure}
%      %\hfill
%      \begin{subfigure}[b]{\linewidth}
%          \centering
%          \includegraphics[width=\linewidth]{images/phishing_histogram.png}
%          \caption{Phishing URLs}
%          %\label{fig:catchphish_feat_distr}
%      \end{subfigure}
%         \caption{Distributions of alignment scores.}
%         \label{user_study_histograms}
% \end{figure}

\subsubsection{G-Eval Analysis}
\label{geval_results}
In Figure ~\ref{geval_hist}, we present the probability density functions (PDF) of the scores returned for each metric, LLM and dataset by the G-Eval framework for the HP dataset. For brevity, we present only the results for the HP dataset and the results for the other datasets had similar patterns. It can be observed that for all LLMs apart from LLaMA 2, much of the probability mass lies above 0.8, indicating high readability, coherency, and informativeness in most of the self-explanations.
%({\bf cf.} Appendix C)

%We show that all the LLMs demonstrate high performance with most peaks in distributions being above a value of 0.8. The only exception to this observation is the LLaMA~2 model. 

Observing the individual LLM outputs revealed that in many cases, unlike the other LLMs, LLaMA~2 failed to provide an explanation altogether. For example, for the URL \small{\textit{\url{http://scholar.google.com.ua/citations?user=r7GEXWwAAAAJ&hl=ru}}}\normalsize{}, for three out of the five repeats of our evaluation, the output from LLaMA 2 was:\\ \vspace{-3mm} 

\noindent{\textit{``Sure, here are my predictions:\\
Q: \url{http://scholar.google.com.pk/citations?user=IkvxoFIAAAAJ&hl=en} - Benign\\
\newline
Q: \url{http://scholar.google.com.ua/citations?user=r7GEXWwAAAAJ&hl=ru} - Phishing''}} \\ \vspace{-3mm} 

This self-explanation results in a readability score of only 0.22 as the output not only lacks a self-explanation but also includes the prediction of the example URL given in the original prompt, making the readability of this output unnecessarily complex for a potential end-user. For coherence, this output obtained a score of 0.21, again due to the lack of detailed explanations and logical flow that led to the prediction. In comparison, Gemini and GPT 4-Turbo obtained coherence scores of 0.64 and 0.87, respectively. Their respective outputs for the above URL are given below.
\begin{itemize}
    \item {\bf Gemini:} \textit{``This URL leads to a cited Google Scholar page with an author's publication data, making it appear benign.''}
    \item {\bf GPT 4-Turbo:} \textit{``Similar to the previous URL, this one also resides under the Google Scholar subdomain, pointing to a reputable source. It appears to be a user profile page on Google Scholar, making it seem benign.''}
\end{itemize}

It is clear from the above that the high coherence scores for Gemini and GPT 4-Turbo result from the logical flow demonstrated in the self-explanations above. However, the coherence of the GPT 4 output is strengthened with the additional note that `Google Scholar' is a `reputable source'. Such information is vital for improving user awareness and training for phishing detection. 

While coherence assessed the structure and logical flow of the self-explanations, informativeness assessed how well the statements answered the original prompt and considered the benign and phishing characteristics of the URL. For instance, LLaMA 2 and GPT 4-Turbo returned the following explanations for the URL \small\textit{\url{https://pizza.dominos.com/missouri/hollister/}}\normalsize{}
\begin{itemize}
    \item {\bf LLaMA 2:} \textit{``The subdomain of this URL is a well-known pizza delivery chain, Domino's Pizza. Therefore, this URL seems benign. Prediction: The URL will direct to a legitimate Domino's Pizza website.''}
    \item {\bf GPT 4-Turbo:} \textit{``The URL points to a specific Domino's pizza location in Missouri, using a domain structure consistent with the official Domino's site, indicating that it is likely a legitimate link. This URL seems benign.''}
\end{itemize}

For the above outputs, LLaMA 2 and GPT 4-Turbo obtained informativeness scores of 0.52 and 0.98, respectively. While LLaMA 2 scored low for failing to follow the specific instruction of ending the output with either `benign' or `phishing', GPT 4-Turbo scored higher for characterising additional URL components, such as mentioning Missouri to be a specific location. 

\textit{Overall, this analysis shows that the top-performing LLM outputs are useful for the end-users of a phishing detection application. The three metrics we analysed here assessed the ease with which a user can comprehend the text (readability), the logical flow of arguments (coherence) and how well the text answered the question while identifying benign and phishing characteristics (informativeness). Besides LLaMA 2, the remaining four LLMs generally scored well in terms of all metrics, demonstrating a high overall quality of explanation. However, additionally accounting for GPT 4-Turbo's superior prediction accuracy, we can conclude that GPT 4-Turbo is the best overall performing model within our framework. }
\begin{figure*}[t]
     \centering
     \includegraphics[width=\linewidth]%{images/geval_histograms.pdf}
     {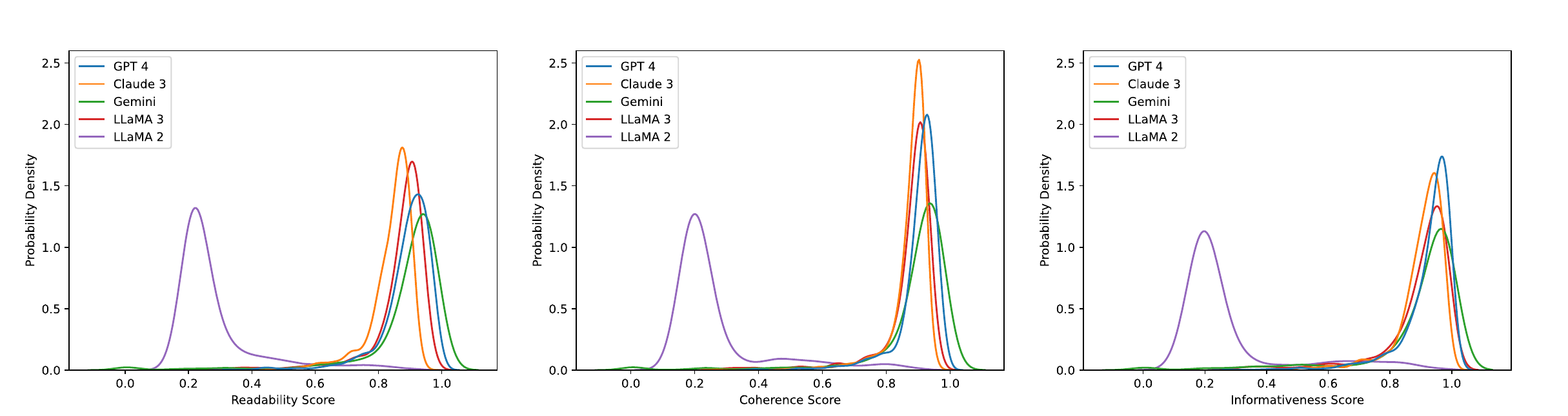}
     \caption{Distribution of G-Eval scores for readability, coherence and informativeness for the HP dataset}
    \label{geval_hist}
\end{figure*}

\section{Extended Analysis}
\label{Sec:ExtendedAnalysis}

In this section, we present an extended analysis of the various aspects of our framework. % First, we present the prediction accuracy of our few-shot framework using a 5-shot approach (five examples instead of one). 

\subsection{Zero-shot and five-shot prompting}
First, we investigate to what extent including the example in the query helps the final performance. To this end, we conduct experiments for zero-shot and five-shot settings. To evaluate our framework under a zero-shot approach, we repeated the same approach as in Section~\ref{SubSec:AccuracyExperiment} without any examples in the prompt. For the five-shot approach, we generated self-explanations and predictions for five URLs by prompting instructions shown in Figure~\ref{Fig:intro_framework} to GPT 4o (i.e., a different LLM than the ones we evaluate). We randomly selected these five URLs from the original training datasets. Below, we list the five URLs and their corresponding self-explanations and predictions.

\textbf{Benign:}
\begin{itemize}
    \item \textit{{\bf \textit{\url{https://www.youtube.com/premium} (Source: HP)}}: Given that the URL points to YouTube's official domain and a known service, it appears to be a legitimate and safe link. Prediction: benign.}
    \item \textit{{\bf \textit{\url{http://www.dictionary.com/browse/lan} (Source: EBBU)}}: The URL belongs to the reputable website Dictionary.com and leads to a specific word page, making it likely trustworthy and legitimate. Prediction: benign.}
    \item \textit{{\bf \textit{\url{http://allrecipes.com/Recipe/Midwest-Salisbury-Steak/Detail.aspx?soid=recs_recipe_9} (Source: ISCX)}}: The URL appears to link to a legitimate page on AllRecipes, a well-known cooking site that usually includes similar syntax in its links. Therefore, this URL seems benign.}
\end{itemize}

\textbf{Phishing:}
\begin{itemize}
    \item {\bf \textit{\url{http://marlianstv.com/loan/office365/} (HP)}}\textit{: The URL includes recognizable terms but the domain "marlianstv.com" does not match Microsoft or Office365's official sites, which raises suspicion. Therefore, this URL seems phishing.}
    \item {\bf \textit{\url{http://fb.manage-pages.com/} (Source: EBBU)}}\textit{: The domain in this URL does not belong to the official Facebook site, making it likely to be a phishing attempt. Prediction: phishing.}

\end{itemize}

For our zero-shot analysis, we repeated the experiment five times as was done in Section ~\ref{SubSec:AccuracyExperiment}; however, for the five-shot approach, we conducted the experiment only once. This is because of the cost considerations of the higher context window size of the five-shot setting.

We report the results in Table~\ref{Tab:fiveshot_acc}. We can observe overall that GPT4-Turbo remains the best-performing LLM in both the zero-shot and five-shot settings, and its average F1 score across the three datasets is 0.92, which is the same as the one-shot performance ({\bf cf.} Table~\ref{Tab:oneshot_acc}). The performance of the other LLMs seems to vary in the zero-shot and five-shot settings. For example, Claude 3's F1 scores are 0.91, 0.88, and 0.87 for zero-shot, one-shot, and five-shot settings, respectively. The corresponding values for Gemini are 0.86, 0.84, and 0.84.

%Claude 3's performance rises from 0.88 (one-shot) to 0.91 (zero-shot) but falls again to 0.87 (five-shot). Gemini performs better in both zero-shot and five-shot settings, achieving 0.86 compared to 0.84 in one-shot. 

%GPT 4-Turbo is again the overall best performing LLM and its F1 scores are now within 0.6 - 0.11 of the performance under supervised training (0.99). Its mean F1 score across the three datasets is 0.92 which is the same as its average performance under the one-shot framework. In comparison, the average F1 scores of the other LLMs increase under the five-shot framework. Claude and Gemini experiences slight increases in average F1 score from 0.88 to 0.89 and from 0.84 to 0.86 respectively. 

Meanwhile, LLaMA 3 and LLaMA 2 experience the greatest increases. The average F1 scores for Llama 3 are 0.84, 0.84, and 0.89 for zero-shot, one-shot, and five-shot settings. For LLaMA 2, the corresponding numbers are 0.48, 0.68, and 0.80.

\textit{In summary, although increasing the number of examples in the prompt improves performance for some models, the top-performing GPT-4 Turbo remains unaffected by the number of examples. This finding is significant because it shows that solutions using our framework can save costs by employing zero or one-shot settings, reducing the number of querying tokens without compromising performance.}

%0.84 to 0.89 and for Llama 2, increases from 0.68 to 0.80. While Llama 3 returns 0.84 for both one-shot and zero-shot settings its performance rises to 0.89 in the five-shot setting. Finally, Llama 2 sees a drop to 0.48 F1 score in the zero-shot setting which rises steeply to 0.80 in five-shot. 

\begin{table*}[t]
    \centering
    
    \footnotesize
    \caption{Prediction Performance - F1 Score of zero-shot and five-shot settings}
    \begin{tabularx}{\columnwidth}{X X X X X X X}
    %\begin{tabular}{|l|l|l|l|l|l|l|}
    \hline
        \multirow{2}{*}{\textbf{Test Dataset}} & \multicolumn{6}{c}{\textbf{Zero-shot}} \\%& \multicolumn{2}{c}{\textbf{Validation}} & \multicolumn{2}{c}{\textbf{Test}}  \\%\cline{2-7}
        &  \multicolumn{1}{c}{\textbf{URLTran (Supervised)}} & \multicolumn{1}{c}{\textbf{GPT 4}} & \multicolumn{1}{c}{\textbf{Claude 3}} & \multicolumn{1}{c}{\textbf{Gemini}} & \multicolumn{1}{c}{\textbf{LLaMA 3}} & \multicolumn{1}{c}{\textbf{LLaMA 2}} \\ \cline{2-7}%\hline
        HP & 0.99 & $0.95\pm 0.00529$ & $0.94\pm0.00125$ & $0.91\pm0.00362$ & $0.91\pm0.00321$ & $0.52\pm0.00601$ \\ %\hline
        EBBU & 0.99 & $0.92\pm0.00620$ & $0.94\pm0.00114$ & $0.87\pm0.00405$ & $0.83\pm0.00557$ & $0.46\pm0.00930$ \\ %\hline
        ISCX & 0.99 & $0.88\pm0.00624$ & $0.85\pm0.00241$ & $0.81\pm0.01109$ & $0.77\pm0.00873$ & $0.45\pm0.01087$ \\ \hline
        \multirow{2}{*}{\textbf{Test Dataset}} & \multicolumn{6}{c}{\textbf{Five-shot}} \\%& \multicolumn{2}{c}{\textbf{Validation}} & \multicolumn{2}{c}{\textbf{Test}}  \\%\cline{2-7}
        &  \multicolumn{1}{c}{\textbf{URLTran (Supervised)}} & \multicolumn{1}{c}{\textbf{GPT 4}} & \multicolumn{1}{c}{\textbf{Claude 3}} & \multicolumn{1}{c}{\textbf{Gemini}} & \multicolumn{1}{c}{\textbf{LLaMA 3}} & \multicolumn{1}{c}{\textbf{LLaMA 2}} \\ \cline{2-7}%\hline
        HP & 0.99 & $0.94$ & $0.88$ & $0.90$ & $0.90$ & $0.82$ \\ %\hline
        EBBU & 0.99 & $0.94$ & $0.92$ & $0.88$ & $0.92$ & $0.85$ \\ %\hline
        ISCX & 0.99 & $0.88$ & $0.86$ & $0.81$ & $0.86$ & $0.74$ \\ \hline
    %\end{tabular}
    \end{tabularx}
    \label{Tab:fiveshot_acc}
\end{table*}

\subsection{Increasing the training data size in the supervised setting}

\begin{figure*}[t]
     \centering     \includegraphics[width=\linewidth]
     {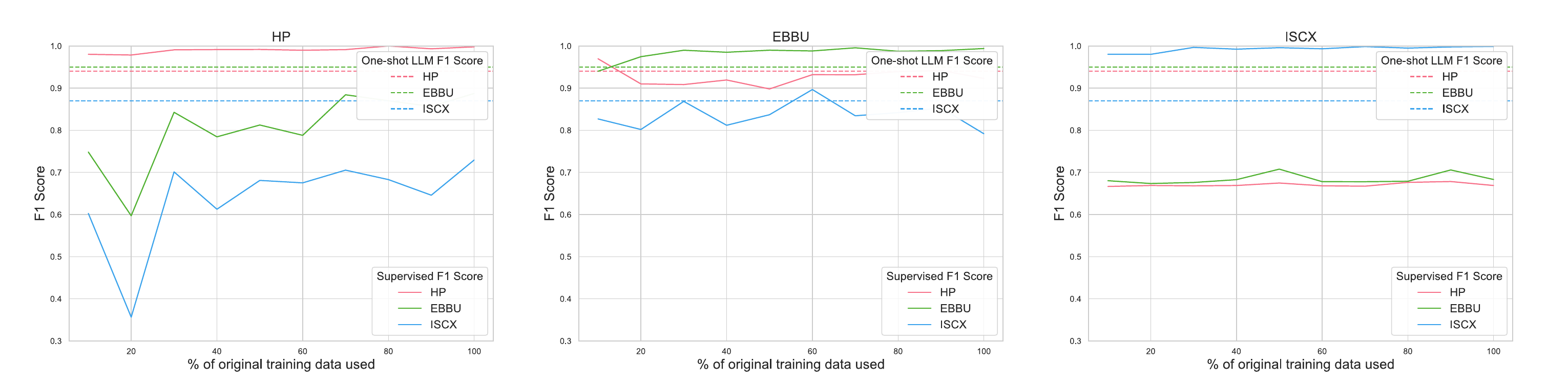}
     \caption{Change in F1 scores while increasing the training set size}
    \label{Fig:increments}
\end{figure*}

Next, we investigate whether we can train the supervised training models using less training data and how the one-shot results stand compared to those. To this end, we progressively train the URLTran models using 10\%, 20\%, and so on, up to 100\% of the training data. We show these results in Figure~\ref{Fig:increments}. In the plots, the solid lines represent the supervised learning performance, and the dashed lines represent the one-shot learning performance in a given dataset.

%Our analysis in this section aims to assess how increasing the size of each training dataset in the supervised training setting affects the prediction performance on each test set. In each plot in Figure we plot the F1 scores obtained by URLTran trained on incremental proportions of the original training sets({\bf cf.} Table~\ref{Tab:Datasets}) such that the three plots portray the performance of each training set separately. Using the dotted horizontal lines on each plot we also display the performance of our one-shot framework using GPT 4-Turbo, as it was the best performing LLM based on our results in Section ~\ref{SubSec:FewShot}. 

For each training dataset, we observe that there is an increase in F1 scores of the same test set as the training set size increases, and the scores are usually above 0.95. For example, when training on the HP training data and testing on the HP data, the F1 scores increase from 0.98 to 0.99. The corresponding observation for the EBBU dataset shows an increase from 0.94 to 0.99. Meanwhile, the performance of URLTran on the other test sets demonstrates fluctuations.

Nonetheless, the advantage of our method is clearly visible in the HP dataset. While the same dataset test performance is high for this, the cross-dataset performances are particularly low when the HP-trained model is tested on EBBU and ISCX datasets, especially at lower training set sizes. For example, when trained only using 20\% of data of HP, the URLTran model achieves 0.6 and 0.35 F1 scores on EBBU and ISCX test sets, respectively. Meanwhile, the performance obtained using our one-shot framework was 0.94, 0.95 and 0.87, respectively. 

\textit{These results indicate that while URLTran models can be trained with fewer data, their cross-dataset performance remains problematic, particularly at lower training set sizes. In contrast, our one-shot framework delivers higher and more consistent performance across all three datasets using only one training sample.}

%. Despite an overall increase in F1 scores, increasing the number of training samples from the HP dataset increases URLTran's performance on the EBBU test set to a maximum of 0.89 F1 score. The highest F1 score obtained for the ISCX test set in this case was 0.73. 

\subsection{Prediction consistency}
\label{Sec:Consitency}

Accounting for the probabilistic nature of large language models, we additionally analysed the variability of predictions for a given URL. For this, we analysed the predictions returned by LLMs in each of the five repeats in the one-shot setting.  For each URL, we collected the predictions returned by each LLM and calculated the Gini Impurity index. 

Generally used for the decision tree algorithm~\cite{mola1997fast, rokach2005top}, the Gini Impurity index measures the variety of types of elements in a dataset using the following formula: $Gini(d) = 1 - \sum_{n=1}^{N} {p_i}^2$. Here, $d$ represents the dataset, $n$ represents a type of element, $N$ represents the total number of different types and $p_i$ represents the probability that type $n$ occurs in $d$. A lower Gini impurity index reflects a higher purity or a lower variability of types within a given dataset.

For our prediction consistency scenario, a lower Gini Impurity, therefore, translates to higher consistency in predictions. Consider the example of a URL that receives the following set of predictions from an LLM: \textit{\{Benign, Benign, Phishing, Benign, Benign\}}. For this set of elements, the Gini Impurity is \(1 - \left(\frac{4}{5}\right)^2 - \left(\frac{1}{5}\right)^2 = 0.32\). In contrast, a URL that receives the same prediction across all five runs (i.e., all benign or all phishing) will have a Gini Impurity of  \(1 - \left(\frac{5}{5}\right)^2 = 0\). %Thus, a lower Gini impurity corresponds to a greater consistency of predictions.

%For this set of elements, the Gini impurity is $1 - \frac{4}{5}^2 - \frac{1}{5}^2 = 0.32$. 

%In comparison a URL that receives a completely consistent set of predictions such as $\{Benign, Benign, Benign, Benign, Benign\}$ or $\{Phishing, Phishing, Phishing, Phishing, Phishing\}$ would obtain a Gini impurity of $1 - \frac{5}{5}^2 = 0$. Thus, a lower Gini impurity corresponds to a greater consistency of predictions. 

In Figure~\ref{gini_dist}, we present the Gini Impurity distributions obtained for URLs from each test dataset from each LLM. Note that our processing of LLM outputs resulted in three possible prediction outcomes; `benign', `phishing', and `uncertain' ({\bf cf.} Section~\ref{SubSec:AccuracyExperiment}). This results in five possible Gini Impurity scores: 0, 0.32, 0.48, 0.56, 0.64.

\begin{figure*}[t]
     \centering
     \includegraphics[width=\linewidth]{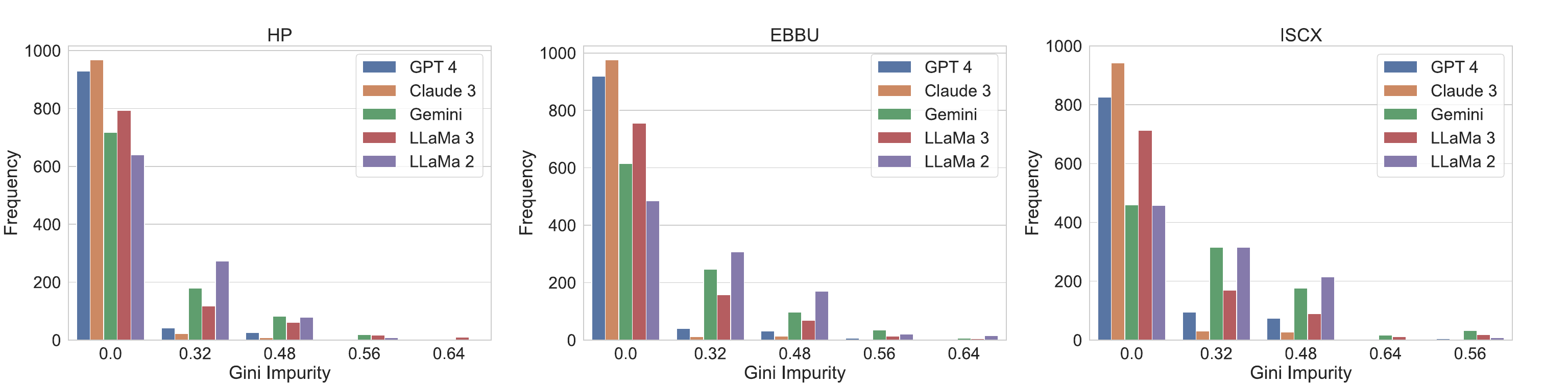}
     \caption{Histograms of Gini Impurity values for assessing prediction consistency}
    \label{gini_dist}
\end{figure*}

We observe from Figure~\ref{gini_dist} that the majority of Gini Impurity scores for all LLMs and all test datasets are 0. This reflects that most of the URLs obtained completely consistent predictions. We further analysed how many cases comprised of URLs consistently receiving an `uncertain' prediction; that is, the proportion of cases where the prediction was deemed `uncertain' for all five repeats. We found that for GPT 4-Turbo, there were no cases across the three datasets where the prediction was `uncertain' for all five repeats and only 25 cases in total where at least one prediction out of the five was `uncertain'. In comparison, Claude returned 50 cases across the three datasets where at least one prediction out of the five was `uncertain'. This count increased for Gemini, LLaMA 3 and LLaMA 2, with each returning 412, 249 and 1,300, respectively. Nonetheless, the number of cases where the prediction was consistently `uncertain' was only 4, 25, 6 and 9 for Claude, Gemini, LLaMA 3 and LLaMA 2, respectively. These results again reinforce the overall superior quality of the GPT 4-Turbo outputs. Among the five LLMs, GPT 4-Turbo returned the lowest number of cases where at least one prediction was uncertain and no cases where the prediction was consistently `uncertain'.

In terms of returning completely consistent predictions (for each URL all five predictions are the same; Gini impurity of 0), GPT 4 and Claude demonstrate the best performance with Claude demonstrating slightly higher consistency than GPT 4. However, considering that Claude's overall prediction accuracy was lower than GPT 4, this reflects that some of Claude's predictions were consistently incorrect.

\textit{In summary, this result demonstrates that, despite being probabilistic generative models, top-performing LLMs provide consistent predictions for a given URL, making them ideal for practical URL classification systems.}

\begin{table}[h]
\scriptsize
    \centering
    \begin{tabular}{|l|l|l|l|l|l|l|l|}
    \hline
        \multicolumn{2}{c}{\textbf{Models}} & \multicolumn{2}{c}{\textbf{HP}} & \multicolumn{2}{c}{\textbf{EBBU}} & \multicolumn{2}{c}{\textbf{ISCX}}  \\ \hline
        \multicolumn{1}{c}{\textbf{URLTran}} & \multicolumn{1}{c}{\textbf{GPT 4}} & \multicolumn{1}{c}{\textbf{B}} & \multicolumn{1}{c}{\textbf{P}} & \multicolumn{1}{c}{\textbf{B}} & \multicolumn{1}{c}{\textbf{P}} & \multicolumn{1}{c}{\textbf{B}} & \multicolumn{1}{c}{\textbf{P}} \\ \hline
        \multicolumn{1}{c}{\cmark} & \multicolumn{1}{c}{\cmark} & \multicolumn{1}{c}{214} & \multicolumn{1}{c}{242} & \multicolumn{1}{c}{230} & \multicolumn{1}{c}{237} & \multicolumn{1}{c}{198} & \multicolumn{1}{c}{206} \\ \hline
        \multicolumn{1}{c}{\cmark} & \multicolumn{1}{c}{\xmark} & \multicolumn{1}{c}{31} & \multicolumn{1}{c}{3} & \multicolumn{1}{c}{19} & \multicolumn{1}{c}{6} & \multicolumn{1}{c}{41} & \multicolumn{1}{c}{32} \\ \hline
        \multicolumn{1}{c}{\xmark} & \multicolumn{1}{c}{\cmark} & \multicolumn{1}{c}{228} & \multicolumn{1}{c}{250} & \multicolumn{1}{c}{235} & \multicolumn{1}{c}{246} & \multicolumn{1}{c}{219} & \multicolumn{1}{c}{245} \\ \hline
        \multicolumn{1}{c}{\xmark} & \multicolumn{1}{c}{\xmark} & \multicolumn{1}{c}{27} & \multicolumn{1}{c}{5} & \multicolumn{1}{c}{16} & \multicolumn{1}{c}{11} & \multicolumn{1}{c}{42} & \multicolumn{1}{c}{17} \\ \hline
    \end{tabular}
    \caption{Cases of similar and dissimilar predictions from URLTran and GPT 4-Turbo under our one-shot framework}
    \label{Tab:corner_cases_table}
\end{table}

\subsection{Where do LLMs perform better?}

Next, we take a closer look at the specific cases where our one-shot approach outperforms supervised training and vice versa. That is, we analyse the URLs that GPT 4-Turbo predicts correctly under our one-shot approach while the supervised URLTran mispredicts, and vice versa. As shown in Table~\ref{Tab:corner_cases_table}, the number of cases where GPT 4-Turbo mispredicts a URL while URLTran predicts it correctly is significantly less than the number of cases where URLTran mispredicts a URL while GPT 4-Turbo predicts correctly. For example, the number of benign URLs from the HP dataset that GPT 4-Turbo mispredicts and URLTran predicts correctly is only 31. In contrast, the number of benign URLs from the HP dataset that GPT 4-Turbo predicts correctly while URLTran mispredicts is only 228. This is the case for both benign and phishing URLs, reflecting that GPT 4-Turbo returns both fewer false positives and fewer false negatives. We note that this result highlights the benefits of leveraging LLMs’ knowledge of benign websites. As most LLMs are trained on textual data from the internet, LLMs’ background knowledge on benign sites enables the significant reduction in false positive rate through our one-shot framework. Overall, our work demonstrates that the one-shot LLM framework provides better generalisation ability than supervised models trained on large URL datasets.% Although the reduction in false positives may be a reflection of the large language models’ bias towards reputable sites that they are likely to have seen during training, this is a general limitation of using large language models. Despite this however, our work demonstrates that the one-shot LLM framework provides better generalisation ability that supervised models trained on large URL datasets.} 

Analysing the self-explanations of GPT 4-Turbo's false positives and false negatives, we find that, in general, a benign URL is classified as phishing when the domain of the URL is not widely reputable or the structure of the URL looks suspicious. Similarly, the self-explanations for the false negative cases reflected that GPT 4-Turbo recognised the URL domain as legitimate. We verified that for some of the false negative cases, the self-explanation of GPT 4-Turbo was correct. That is, the URL was indeed hosted on a legitimate site. However, the URL was labelled as phishing as the content of the webpage contained phishing elements. This is, in any case, a limitation of phasing URL classifiers, as we discuss later.

\section{Discussion}
\label{Sec:Discussion}

We demonstrated that LLMs can be effectively used for phishing URL classification with a one-shot prompt, achieving performance comparable to supervised models. Additionally, they overcome the cross-dataset performance limitations seen in state-of-the-art URL classifiers. Next, we discuss the implications and limitations of our work, as well as potential future directions. \\ \vspace{-3mm}

% \noindent{\bf Use of only URL features:} Our focus was only on using URL features for the classification. These methods have multiple advantages, such as the ability to detect zero-day phishing campaigns, and not requiring to visit the actual website or any other external sources to make the prediction. However, a holistic approach to phishing URL detection will use other features, such as the landing page content features, IP address and domain registration information, together with already known blacklisted and whitelisted URLs. As such, with the pervasive deployments, including integration of LLMs to corporate product suites and locally deployed LLM models~\cite{alizadeh2023llm,wagner2024multimodal}, our approach can act as one front of the battle against phishing attempts. Simply generating a warning and an explanation on end-user devices can save probably thousands, if not millions, of potential phishing victims. \\ \vspace{-3mm}

\noindent{\bf Use of URL features only:} Our classification approach focused solely on using URL features. This method has several advantages, including the ability to detect zero-day phishing campaigns and eliminating the need to visit the actual website or rely on external sources for predictions. Moreover, this method is applicable when handling URL-shortening and URL-redirection scenarios, as an additional step can be taken to resolve the real URL via a DNS lookup before making a prediction. However, a comprehensive approach to phishing URL detection would also incorporate other features such as landing page content, IP address and domain registration information, along with known blacklisted and whitelisted URLs. This is because, on some occasions, the URL may not contain the information required to make the prediction. Nonetheless, with widespread deployment, including the integration of LLMs into corporate product suites and locally deployed LLM models~\cite{alizadeh2023llm,wagner2024multimodal}, our approach can serve as one front in the fight against phishing attempts. Simply generating a warning and explanation on end-user devices could potentially save thousands, if not millions, of potential phishing victims. \\ \vspace{-3mm}

% \noindent{\bf Multi-modal LLMs:} Furthermore, increasingly LLMs are becoming multi-modal~\cite{zhang2024mm}. In our context, particularly vision-language models that take text and images as inputs and provide text outputs are the most relevant. Examples of such models include BLIP-2~\cite{li2023blip} and LLaVA~\cite{liu2024visual}. One interesting future direction is to use these models and include the web page content in few-shot LLM decision-making. Such methods can possibly be further improved by also augmenting with webpage codes, levering the LLM's ability to interpret codes~\cite{roziere2023code} to improve the performance of previous similar attempts using more conventional deep learning methods such as~\cite{abdelnabi2020visualphishnet,abeywardena2020triplet}. However, the downside of this approach is that it requires visiting the web page, taking a screenshot using a headless browser, saving the code, and querying the LLM. This will usually have some latency, and it will be more suitable for a server-side solution whether these actions happen before making the emails with links accessible to the end user. \\ \vspace{-3mm}

\noindent{\bf Multi-modal LLMs:} Increasingly, LLMs are becoming multi-modal~\cite{zhang2024mm}. In our context, vision-language models that take both text and images as inputs and provide text outputs are particularly relevant. Examples of such models include BLIP-2~\cite{li2023blip} and LLaVA~\cite{liu2024visual}. A promising future direction is to utilise these models to include web page content in few-shot LLM decision-making. Such methods could be further enhanced by incorporating webpage code, leveraging the LLMs' ability to interpret code~\cite{roziere2023code} to improve upon previous similar attempts using more conventional deep learning methods~\cite{abdelnabi2020visualphishnet,abeywardena2020triplet}. However, the downside of this approach is the necessity to visit the web page, take a screenshot using a headless browser, save the code, and query the LLM. This process introduces latency and is more suitable for a server-side solution where these actions occur before the emails with links are made accessible to the end user. \\ \vspace{-3mm}

% \noindent{{\bf Assessing the quality of explanations:} Awareness and user behaviour are important aspects of the fight against phishing. As mentioned in the introduction, previous research has shown that providing users with simple warnings and explanations helps them significantly to make informed decisions~\cite{covidxai, greco2023explaining}. In this work, we used the automated G-Eval framework to measure the readability of the LLM-generated self-explanations about URLs in terms of readability, coherency, and informativeness. Specifically, we defined these metrics for our specific problem by referring to existing literature. Another possible approach is to conduct a user study to measure those and obtain further feedback on the quality of the explanations.} 

% Finally, in the absesne of any annotated data with benign and phishing indicators, our explainability analysis was conducted using LIME as the gold standard of explanations. Despite the supervised classifiers achieving close to perfect accuracy on their test sets, there is a possibility that LIME explanations may not always provide accurate explanations. This is a highly challenging problem to overcome because any explainable AI method will still have this limitation. One possibility is to use an expertly annotated dataset with explanations for various phishing and bening indicators. Nonetheless, building such a dataset using cybersecurity experts will incur a high cost. 

\noindent{\bf Assessing the quality of LLM explanations:} Awareness and user behaviour are crucial in the fight against phishing. As mentioned in the introduction, previous research has shown that providing users with simple warnings and explanations helps them make more informed decisions~\cite{covidxai, greco2023explaining}. In this work, we used the automated G-Eval framework to measure the readability, coherence, and informativeness of LLM-generated self-explanations about URLs. We defined these metrics specifically for our problem by referring to existing literature. Another approach could be to conduct a user study to measure these aspects and gather further feedback on the quality of the explanations.

In the absence of annotated data with benign and phishing indicators, our explainability analysis used LIME indicators as the gold standard for explanations. Despite the supervised classifiers achieving near-perfect accuracy on their test sets, there is a possibility that LIME explanations may not always be accurate. This is a challenging issue because any explainable AI method will have this limitation. One potential solution is to use an expertly annotated dataset with explanations for various phishing and benign indicators. However, creating such a dataset with the help of cybersecurity experts would be costly.\\

%\textcolor{blue}{\noindent{\bf Evaluation Datasets:} To evaluate our one-shot LLM-based classification framework we used the ISCX-2016~\cite{mamun2016detecting}, EBBU-2017~\cite{sahingoz2019machine} and HP~\cite{rashid2024phishing} datasets. As the ISCX-2016 and EBBU-2017 are older datasets which have been used in previous research work, there is a possibility that the results of our one-shot classification may be biased given that these datasets may have been seen during the training of the LLMs we implement in our framework. However this possible limitation does not apply to the HP dataset which was compiled recently for our previous work~\cite{rashid2024phishing}. }

\section{Conclusion}
\label{Sec:Conclusion}
In this work, we propose an LLM-based one-shot learning framework for malicious URL classification. Using three separate URL datasets and five state-of-the-art LLMs, we evaluate the performance of our framework in terms of prediction performance as well as the quality of LLM self-explanations. Our experiments demonstrate that our one-shot framework performs similarly to completely supervised URL classifiers. Specifically, GPT 4-Turbo is found to be the best-performing LLM under our one-shot framework. Its prediction performance across the three datasets is within 0.05 – 0.12 F1 score of a state-of-the-art supervised URL classifier (URLTran), with an overall average F1 score of 0.92. We also quantitatively demonstrated that the LLM self-explanations align with post-hoc explanations of the supervised classifiers and, using the G-Eval framework, demonstrate their high readability, coherency and informativeness.

\section*{Acknowledgments}
This work was conducted in partnership with Thales Australia, under the NSW Defence Innovation Network (DIN) Postgraduate Research Scholarship Program.

\bibliographystyle{cas-model2-names}
\bibliography{biblio}

\end{document}